# A Cluster-Based Opposition Differential Evolution Algorithm Boosted by a Local Search for ECG Signal Classification

Mehran Pourvahab[*a,b], Seyed Jalaleddin Mousavirad[a], Virginie Felizardo[a,b],
Nuno Pombo[a,b], Henriques Zacarias[a,b,e], Hamzeh Mohammadigheymasi[a],
Sebastião Pais[a,d,g], Seyed Nooreddin Jafari[a,f], Nuno M.Garcia[b,c]

[a] Department of Computer Science, University of Beira Interior, Covilhã, Portugal
[b] Instituto de telecomunicações, Covilhã, Portugal
[c] Physics Department, Faculty of Sciences, University of Lisbon, Portugal
[d] NOVA LINCS, New University of Lisboa, Lisboa, Portugal
[e] Polytechnic Institute of Huila, Mandume Ya Ndemufayo University, Lubango, Angola
[f] Department of Electrical Engineering, Langarud Branch, Islamic Azad University, Langarud, Iran
[g] Groupe de Recherche en Informatique, GREYC, University of Caen Normandie, Caen, France

## Abstract

Electrocardiogram (ECG) signals, which capture the heart's electrical activity, are used to diagnose and monitor cardiac problems. The accurate classification of ECG signals, particularly for distinguishing among various types of arrhythmias and myocardial infarctions, is crucial for the early detection and treatment of heart-related diseases. This paper proposes a novel approach based on an improved differential evolution (DE) algorithm for ECG signal classification for enhancing the performance. In the initial stages of our approach, the preprocessing step is followed by the extraction of several significant features from the ECG signals. These extracted features are then provided as inputs to an enhanced multi-layer perceptron (MLP). While MLPs are still widely used for ECG signal classification, using gradient-based training methods, the most widely used algorithm for the training process, has significant disadvantages, such as the possibility of being stuck in local optimums. This paper employs an enhanced differential evolution (DE) algorithm for the training process as one of the most effective population-based algorithms. To this end, we improved DE based on a clustering-based strategy, opposition-based learning, and a local search. Clustering-based strategies can act as crossover operators, while the goal of the opposition operator is to improve the exploration of the DE algorithm. The weights and biases found by the improved DE algorithm are then fed into six gradient-based local search algorithms. In other words, the weights found by the DE are employed as an initialization point. Therefore, we introduced six different algorithms for the training process (in terms of different local search algorithms). In an extensive set of experiments, we showed that our proposed training algorithm could provide better results than the conventional training algorithms.

Keywords: Neural networks, differential evolution, regularization, clustering, opposition-based learning, ECG analysis

---

[*] Corresponding author.
  E-mail address: mehran.pourvahab@ubi.pt

## 1. Introduction

Heart diseases, also known as Cardiovascular Diseases (CVDs), are the leading cause of death, accounting for approximately 32 percent of all deaths. Around 85 percent of these deaths were attributable to heart attacks and strokes. The heart is a cone-shaped muscular organ that contracts at regular intervals to supply blood to the body's organs [1–5]. CVDs result in irregular heartbeats called arrhythmia, and sudden death can occur depending on the severity of the arrhythmia. In addition, a heart attack is caused by a blockage in the coronary arteries, which supply blood and oxygen to the heart. Diet, hypertension, smoking, and other changes in lifestyle are the primary causes of cardiovascular diseases. According to statistics from the World Health Organization (WHO) [6], an estimated 17.9 million people died from cardiovascular diseases in 2019. Moreover, the majority of these deaths occur in low- and middle-income nations[7]. Recent years have seen an increase in the implementation of various programs and policies aimed at preventing and reducing the occurrence of first and recurring cardiovascular events in increasingly diverse communities[8]. The Electrocardiogram (ECG/EKG) has emerged as the most popular bio-signal for the early detection of CVDs in order to meet this objective.

The ECG graphically depicts the electrical activity of the human heart, and the signal morphologies of the ECG provide information about different forms of arrhythmia depending on specific cardiac diseases. The ability to quickly and accurately diagnose arrhythmia, other cardiac conditions, and anomalies from the ECG wave graph has the potential to save countless lives and significantly reduce the cost of healthcare globally [9]. The ECG signals are exceedingly challenging to interpret due to the nature of non-stationary. These types of signals are non-stationary and require a lengthy period of clinical monitoring. As a result, ECG analysis uses a computer-based approach. Electrical activity is the ECG signal's underlying concept. The electrical activity spreads throughout the body and is absorbed by the skin. Finally, the electrodes on the ECG machine capture the activity and show it visually [10].

Machine learning algorithms have emerged as promising alternatives to traditional methods for diagnosing heart diseases from ECG signals [11]. Similarly, in other medical signals like EEG, these algorithms show potential [12,13]. In recent years, numerous studies have explored the potential of artificial intelligence (AI) techniques for interpreting ECG signals and diagnosing cardiovascular diseases [14–16]. However, there are several machine learning algorithms; there are stills many open doors for enhancing the effectiveness of the prediction results. In light of these challenges, this study aims to apply a novel machine learning method on ECG signals for classification under MLP, boosted by an innovative approach to optimization, the Differential Evolution (DE) algorithm, which has been developed. Our proposed differential evolution algorithm uses a region-based strategy and opposition-based learning to find the optimal weights and biases in the employed MLP. To ameliorate the DE-based trainer, we also fed the weights yielded by the proposed DE as an initialization step to a local search.

Artificial neural networks (ANN) employ feed-forward neural networks (FFNN) as a common topology to tackle intricate classification and regression problems. Neurons and their connections make up the basic components that form FFNN. The data inputs in FFNNs are directed in a single flow and traverse through intermediary levels, which are called the hidden layers, before arriving at the output layer. The multilayer perceptron (MLP), in which neurons are organized cascade-wise, is the most widely used FFNN model. At least two layers make up MLP. In MLPs, no information transfer between the neurons makes up a layer; instead, the inputs to the neurons of the $i+1$-th layer are the $i$ −th layer's outputs. The input layer and output layer are the first and last layers, respectively, while the hidden layer is the layer that lies between the input layer and the output layer and contains several processing nodes. The number of output classes is represented by the number of nodes in the output layer, whereas the number of nodes in the input layer corresponds to the number of features in the input vector.

However, depending on the task, the hidden layer's node count changes and is best determined by experimentation and trial-and-error [17]. Each link in an FFNN has one weight, which represents its strength. Finding suitable weights that lower the error between the actual and predicted outputs is the goal of training in FFNNs. Gradient-based methods, such as the back-propagation algorithm, are so prevalent in the literature, while they suffer from some drawbacks such as getting stuck in a local optimum and sensitivity to the initialization point [18].

In the search for dependable alternatives to traditional algorithms, Population-based Metaheuristic (PBMH) algorithms such as Particle Swarm Optimization (PSO) [19] and Differential Evolution (DE) [20] stand out as feasible options. Within the realm of PBMH, Evolutionary Algorithms (EA) have been widely utilized for Multi-Layer Perceptron (MLP) training and have demonstrated superior effectiveness, efficiency, and user-friendliness when compared to Back-Propagation (BP) and Genetic Algorithm (GA) for MLP training [21]. As a matter of fact, GA has been found to excel in this regard. Researchers have even developed a GA-based method that is specifically tailored for rapid MLP training, which has shown to be more efficient than conventional GA-based training algorithms that use a combination of GA and BP for MLP weight determination [22]. To that end, this technique has been proven to outperform both GA and BP [23].

Swarm intelligence algorithms [24] are another category of PBMHs that inspired the swarm behavior in the nature. A hybridized approach has been proposed by researchers that incorporate accelerated PSO with Levenberg-Marquardt to enhance the convergence rate of PSO-based training for medical datasets [25]. Through extensive experimentation on multiple clinical datasets, the efficacy of this approach has been confirmed. Other techniques that have been utilized for training multilayer perceptrons (MLPs) include but are not limited to artificial bee colony [26], grey wolf optimizer (GWO) [27], imperialist competitive algorithm [28], firefly algorithm [29], whale optimization algorithm (WOA) [30], ant lion optimizer, the dragonfly algorithm (DA) [31], sine cosine algorithm [32], grasshopper optimization algorithm, and salp swarm algorithm (SSA) [33].

Differential Evolution (DE) [20] is a well-established and efficient PBMH that has demonstrated outstanding performance in addressing complicated optimization issues [34,35], and numerous studies have been conducted in recent years to improve the DE algorithm [36]. The process involves three main operators: mutation, crossover, and selection. Mutation gathers data from multiple potential solutions, while crossover combines the mutated and target vectors. The selection operator picks the best candidate from both old and new solutions to make up the next population.

The DE algorithm is proposed for MLP training, and it is indicated superior performance compared to gradient-based techniques [37]. In another study, [38] employed opposition-based learning in conjunction with DE (QODE) and showed that QODE outperforms its competitors on a variety of classification challenges. In one of the most recent publications, the authors deployed a DE algorithm enhanced by opposition-based learning and a region-based technique (RDE-OP) [39].

This paper presents a novel DE-based trainer for ECG classification problems, incorporating opposition concepts, a centroid-based scheme, and a local search. This approach, combining the strengths of multiple algorithms and techniques, offers a promising solution to ECG signal classification problems with potential applications in clinical and research settings.

Our ECG classification approach aims to differentiate between 'Normal' and 'Not Normal' ECG signals. We have utilized the PTB-XL [40] dataset, a comprehensive resource for ECG signal studies. To streamline our classification approach and align with our primary objective, we transformed the superclass labels of the PTB-XL dataset into a binary format. The 'Normal' ECG signals are labeled as '1', representing typical heart rhythms, while all other types of signals, indicative of any form of irregularity or abnormality, are labeled as '0'. This binary classification plays a crucial role in our methodology and simplifies the process of ECG signal classification.

The main characteristics of our proposed approach are as follows:

- We propose a novel ECG signal classification algorithm that combines an MLP with an improved DE algorithm and a gradient-based local search algorithm for the training process, resulting in an effective and efficient classification method.

- The boosted DE method uses a clustering strategy as a multi-parent crossover operator to increase its capacity to exploit data and promote better convergence.

- The improved DE algorithm incorporates quasi-based opposition-based learning to boost exploration capabilities and significantly facilitate the search for optimal solutions.

- The ideal weights discovered by the improved DE technique are utilized to initiate the gradient-based local search, resulting in more precise and effective model fine-tuning.

- Six different local search algorithms are used, producing six distinct approaches for classifying ECGs, allowing flexibility and adaptation to various issue situations.

- Our proposed ECG signal classification method is thoroughly assessed using cutting-edge benchmarks and extensive experiments, proving its efficacy and superiority over current approaches.

The rest of the paper is organized as follows. Section 2 reviews ECG, neural networks, pattern clustering, and differential evolution. Section 3 introduces the proposed algorithm, followed by an explanation of the general structure. The data used in the research, including preprocessing data, MLP boosted by an improved DE algorithm and Local Search, are presented in this section. In Section 4, the proposed approach is evaluated and discussed by several criteria. Lastly, Section 5 will conclude the paper.

## 2. Background Knowledge

This section provides an overview of the background knowledge pertinent to our research objectives, covering key aspects such as electrocardiogram (ECG) signals, neural networks, pattern clustering, and differential evolution. A comprehensive understanding of these concepts is essential for the development and evaluation of our proposed ECG signal classification method. We will delve into the characteristics of ECG signals, including various waveform components and intervals, followed by an examination of the neural networks' principles and applications.

Next, we will discuss pattern clustering techniques and their relevance to our approach. Finally, we will explore the concept of differential evolution, a powerful optimization technique that plays a crucial role in the training process of our proposed method.

### 2.1. ECG

A Dutch doctor and physiologist, Willem Einthoven, came up with the idea of electrocardiography more than a century ago. Figure 1 shows the traditional ECG curve and some of its most frequent waveforms. The diagram shows significant time intervals and measurement sites. Understanding these waves and intervals is necessary for ECG interpretation.

Electrodes on the ECG leads pick up the tiny potentials created by the heart's tissues, which is how the electrical activity is measured. The ECG is created by amplifying the tiny impulses and storing them in a database. It is compulsory to be familiar with these waves and intervals to interpret an ECG. P, Q, R, S, and T are the five distinct peaks and valleys that may be seen in an ECG signal. Another peak termed U is also used in certain circumstances. The PR, QT, RR, and QRS intervals on an ECG signal are three different intervals. In the following, we describe the terms in detail.

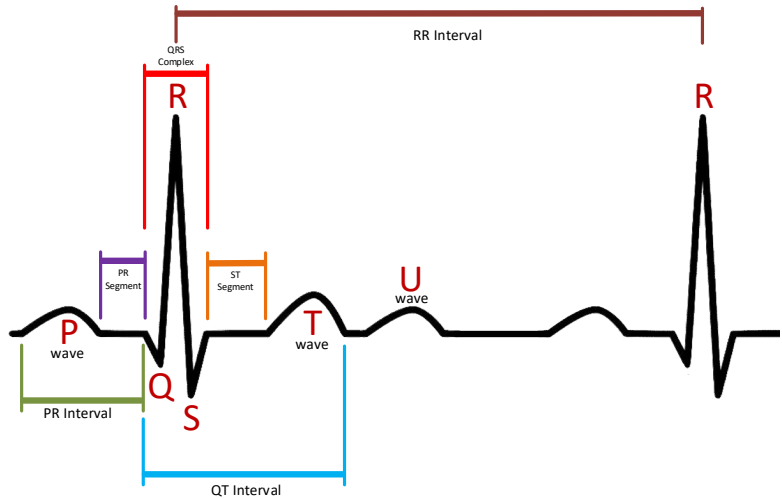

Figure 1: Electrocardiogram (ECG) waveforms

### 2.1.1. The P-Wave, PR interval, and PR segment

P-wave evaluation is generally the first step in ECG interpretation. An atrial depolarization (P-wave) is shown by the P-wave (activation). An electrocardiogram's PR interval measures how far apart the P-wave and QRS complex seem to be. The PR interval is measured to see whether the atrioventricular impulse conduction is normal. It is termed the PR segment, and it shows the delayed conduction of the atrioventricular node's impulses between the end of the P-wave and the commencement of the QRS complex. Normal PR Interval values range between 0.12 and 0.20 seconds. As a reference line or isoelectric line, it is referred to as the PR segment (sometimes referred to as the reference line). The PR segment is used as a baseline to quantify the amplitude of any deflection or wave.

### 2.1.2. The QRS complex

When it comes to interpreting an ECG, the QRS complex is the most critical waveform because it depicts the heart's electrical activity during ventricular contraction, which is a key indicator of the heart's health [41,42]. Ventricle depolarization (activation) is represented by the QRS complex. QRS complex, even if all three waves are not always visible, is usually referred to as such. In cardiac electrophysiology, the QRS complex's duration is measured in milliseconds. Having a brief QRS complex indicates that the ventricles depolarize quickly, which in turn indicates that the conduction system is working effectively. It is possible that the delayed rate of ventricular depolarization seen in patients with wide QRS complexes is the result of conduction system malfunction. The average range of the QRS complex is less than 0.10 seconds. In addition, between 0.5 and 0.7 mV is the usual range for QRS amplitude. ECG signal includes certain noises.

### 2.1.3. ST Segment

The plateau phase of the action potential is represented by the ST section. Since it is changed under a broad variety of circumstances, the ST segment should constantly be carefully examined. Several of these diseases result in ST segment alterations that are very distinctive. Acute myocardial ischemia necessitates close attention to the ST segment because of the ST segment's tendency to deviate (ST segment deviation).

### 2.1.4. T Wave

Repolarization of the ventricles is represented by the T wave. There are several scenarios in which T-wave alterations occur, and the T-wave represents the repolarization of contractile cells. T-wave transitions should be easy to navigate and not abrupt. The regular T-wave has a greater downward slope and is somewhat asymmetrical.

### 2.1.5. U Wave

The U-wave is infrequently seen in ECG signals. Its amplitude is typically one-fourth that of the T-wave. It is a positive wave that follows the T-wave. Candidate solutions with significant T-waves and sluggish heart rates exhibit U-waves more often, and the origins of the U-wave remain obscure.

*2.1.6. QT Interval*

The length of the electrical systole is represented by the QT interval. It begins at the beginning of the QRS complex and concludes at the conclusion of the T wave. Less than 0.42 seconds is the QT interval's average value. In addition, the QT interval rises with slower heart rates and decreases at higher heart rates, demonstrating the inverse relationship between the QT length and heart rate. Therefore, it is vital to account for the heart rate in order to evaluate if the QT interval is within normal bounds.

*2.1.7. RR Interval*

The RR interval is the period between successive heartbeats or the distance between two consecutive heartbeats that indicted to interval refers to the time between the commencement of one R wave and the onset of the next R wave, representing one complete cardiac cycle. Also, the RR interval is the essential ECG characteristic for diagnosing heart illness. Furthermore, the RR interval calculates heart rate variability (HRV) characteristics [43]. Analysis of heart rate variability is a noninvasive measurement representing the variation in time between successive heartbeats. The HRV characteristics are time domain and frequency domain characteristics. The mean and standard deviation of the RR interval are calculated using time domain analysis [44].

*2.2. Neural Networks*

As a well-known family of artificial neural networks (ANNs), feedforward neural networks (FFNN) are a technique for supervised pattern recognition with widespread applicability [45,46]. Among these ANNs, feedforward neural networks (FFNN) and multilayer perceptrons (MLP) are prominent techniques. FFNNs consist of three primary layers: an input layer, an output layer, and hidden layers. Each node within these layers contains an activation function that describes how the output should be affected by the weighted sum of the inputs. One of the most popular activation functions is the Sigmoid function, which is defined as:

$$\delta(net) = \frac{1}{1 + e^{-net}} \tag{1}$$

where $net$ is the input variable. In a Feedforward Neural Network (FFNN), the connection between layers is given a weight representing the strength between two nodes. These weights are essential to the performance of the FFNN. However, the process of identifying optimal weight values, commonly referred to as training, is widely regarded as a challenging undertaking. The most common approach for training FFNNs is through Gradient Descent-based techniques (GD).

Multilayer perceptron (MLP) is a type of feedforward neural network consisting of multiple layers of interconnected nodes, including input, hidden, and output layers. The MLP learns to map input patterns to corresponding output patterns through a supervised learning process, adjusting the weights of the network to minimize the error between the predicted and actual outputs. Figure 2 demonstrates an example of an MLP, which is conceptually similar to the FFNN but typically has more hidden layers and nodes. Like the FFNN, the MLP uses activation functions, such as the Sigmoid function, to modulate the output of each node based on the weighted sum of its inputs, as shown in Equation (1). The primary difference between the two architectures lies in the complexity of the MLP, which allows it to learn more sophisticated patterns and representations in the data, leading to improved performance in many applications. Training an MLP involves adjusting the weights using gradient descent-based techniques or other optimization algorithms to minimize the error between the predicted and actual outputs.

### 2.3. Pattern Clustering

The goal of clustering is to organize patterns into groups called clusters. This is accomplished by reducing the distances between samples that are contained within the same cluster, known as intra-cluster distances, while simultaneously increasing the distances that separate samples from different clusters, known as inter-cluster distances.

Assume a collection of $N$ patterns with the notation $P = \{p_1, p_2, \ldots, p_N\}$ and $p_i = \{p_{i,1}, p_{i,2}, \ldots, p_{i,d}\}$, where d is the problem dimension. The goal of clustering is to identify $K$ partitions $C = \{c_1, c_2, \ldots, c_K\}$ such that

$$c_i \neq \emptyset, i - 1 \ldots K \tag{2}$$

$$\cup_{i=1} c_i = P \tag{3}$$

$$c_i \cap c_j = \varphi, i, j = 1 \ldots K, i \neq j \tag{4}$$

Typically, a similarity criterion is used to group samples into clusters. In this study, the applied similarity criterion is the Euclidean distance.

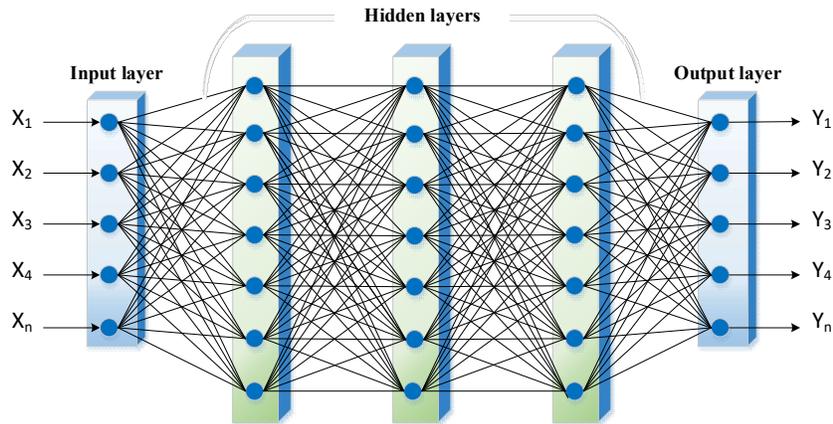

Figure 2: A MLP with hidden layer structure.

### 2.4. Differential Evolution

The utilization of Differential Evolution (DE) [20] as a population-based metaheuristic has proven to be a viable solution for addressing complex optimization problems. DE uses mutation, crossover, and selection as its three main operations and starts with NP candidate solutions created randomly and taken from a uniform distribution. A mutant vector is produced by mutation, one of the most well-known approaches, as

$$v_{i=(v_{i,1}, v_{i,2}, \dots, v_{i,D})}$$
(5)

which is generated as

$$v_{i=x_{r1}+F*(x_{r2}-x_{r3})}$$
(6)

where $F$ is the scale factor and $x_{r1}$, $x_{r2}$ and $x_{r3}$ are three unique candidate solutions that are randomly chosen from the present population. The mutant vector is integrated into the target vector through the crossover operator. The technique utilized in this work is called binomial crossover, which can be explained as follows:

$$u_{i,j=} \begin{cases} v_{i,j} & rand(0,1) \leq CR \ or \ j == j_{rand} \\ x_{i,j} & otherwise \end{cases}$$
(7)

where $i = 1, \dots, NP, j = 1, \dots, D, CR$, and $j_{rand}$ is a random number between 1 and $N_P$. Also, CR is the crossover rate. In the DE algorithm, the selection operator is responsible for choosing the most suitable candidate solution from the current and previous options. The processing of training using a population-based approach is shown in Figure 3.

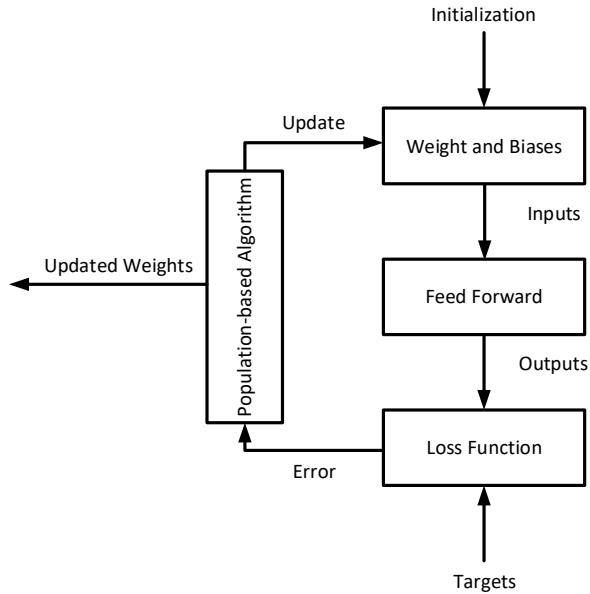

Figure 3: Training Process of the MLP using Population-Based algorithm.

## 3. Methodology

This study proposes a brand-new training procedure for ECG signal classification called CODEL for determining an MLP's optimal weights. To this end, first, we employ a real-valued representation to encode the weights and biases. Then, we define an objective function based on the classification error. For the search strategy, we used an improved DE algorithm as a global search, so that it integrates the main mechanisms of the DE algorithm with opposition-based learning and a region-based strategy. The weights and biases found from the previous step are fed to a local search. In other words, the weights found by the improved algorithm are responsible for initialization for the local search. The general structure of our proposed algorithm is indicated in Figure 4, while further details are provided in the following.

As depicted in Figure 4, our approach comprises multiple stages of operation. As shown in the figure, the input ECG signals are preprocessed in a variety of methods to increase signal quality. Afterwards, it is served as input to a transformer to transform it into a set of features. The normalized data were then utilized as inputs for the improved neural networks. The overall goal of the proposed CODEL methodology is to distinguish between 'Normal' and 'Not Normal' ECG signals in a binary classification format, thus aiding in the detection of irregular or abnormal heart rhythms.

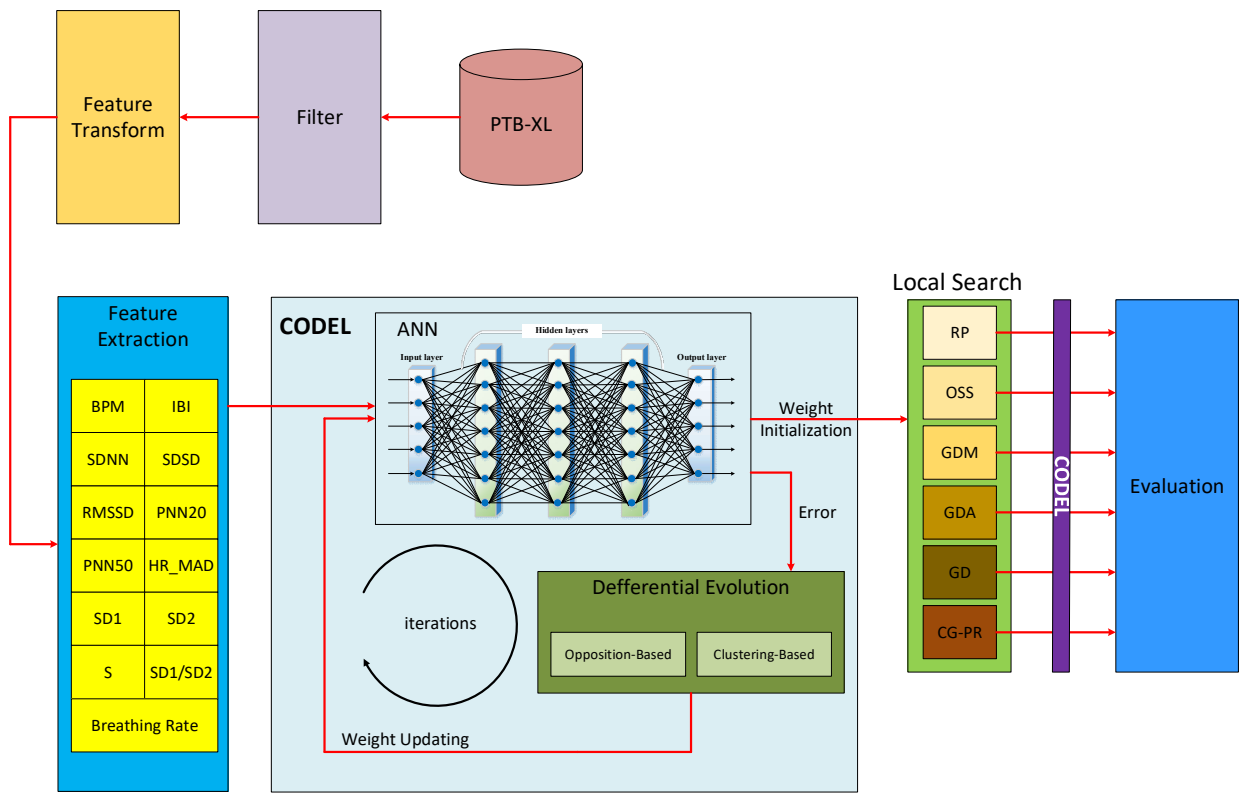

Figure 4: Working Process of CODEL Method.

*3.1. DATA*

The PTB-XL dataset, which is the basis for experimental results provided in this research, is briefly introduced in this section. The PTB-XL dataset, the most extensive publicly accessible electrocardiography dataset, is used in this study [47]. It offers a comprehensive collection of ECG annotations, encompassing a broad range of heart conditions such as various types of arrhythmias and myocardial infarctions, thereby enabling the development, and testing of machine learning algorithms designed to classify these conditions. It comprises 21,837 combined recordings of 10-s long 12-lead electrocardiograms from 18,885 different individuals, of which 52% were men and 48% were women. The PTB-XL dataset is balanced between genders, as mentioned earlier in the percentage of genders, ranging in age from two to ninety-five (median 62). The dataset was augmented with additional patient information (age, sex, height, weight). ECG files are also available in two different variations with sampling speeds of 500 Hz and 100 Hz with 16-bit resolution. Up to two cardiologists annotated the ECG recordings. The SCP-ECG standard is met by several ECG records from 71 different recordings [48]. In conclusion, PTB-XL stands out not just as the most significant publicly available clinical ECG dataset to date but also for its comprehensive collection of ECG annotations and other information, which makes it a perfect resource for training and evaluating machine learning algorithms. Throughout this study, input data with a sampling frequency of 100 Hz are used. For the purpose of this study, we transformed the superclass labels of the PTB-XL dataset into a binary classification format. The ECG signals that were classified as "Normal" were labeled as '1', representing standard heart rhythms. In contrast, all other types of classified, which indicate any form of irregularity or abnormality, were labeled as '0'. This binary classification simplifies the ECG signal classification process and aligns with our main objective of differentiating between typical and atypical heart rhythms.

*3.2. Preprocessing*

Before proceeding with feature extraction and model development, preprocessing the ECG signal is a crucial step. To this end, the dataset is first cleaned, and the ECG signal is standardized to have zero mean and unit variance. Although most signals in the dataset are of high quality, some may contain artifacts such as powerline noise, baseline drift, burst noise, and static noise. Therefore, it is vital to remove these artifacts before applying machine learning models to ensure accurate analysis and prediction. In order to address high-frequency noise in the ECG signal, a Butterworth filter is employed. Butterworth filters, also known as "maximally flat magnitude filters," are a type of signal processing filter specifically designed to provide the flattest possible passband frequency response [49]. This characteristic ensures minimal distortion of the desired signal components while effectively removing noise. In addition to filtering, outlier detection is implemented on the raw signal using a modified Hampel Filter [50] with a window size set to half the sampling rate. The Hampel Filter is a reliable method that can help identify and reduce the impact of outliers in data. This can significantly enhance the accuracy and trustworthiness of any subsequent analysis. The preprocessing stage leverages the Heartpy toolkit [51,52], an open-source Python library for ECG signal processing and analysis. Some modifications have been made to the toolkit better to suit the specific requirements of the current study, ensuring optimal performance and compatibility with the dataset. By fully preprocessing the ECG signals, the quality of the input data is increased, and the possibility of artifacts interfering with the analysis is minimized. This process lays the foundation for accurate feature extraction and effective model development, ultimately leading to more reliable predictions and insights into underlying physiological processes.

### 3.3. Feature Extraction

The ECG signal needs to be comprehensively represented by a criterion choice of features, either based on statistical parameters or on dynamic characteristics, also keeping in mind that the selected features will need to be combined with the raw ECG signal to form the feature set [53]. Common examples of statistical features are maximum, minimum, mean, standard deviation, kurtosis, and skewness. The statistics allow a birds-eye view of the signal waveform, including its amplitude, dispersion, and symmetry. Still, they are oblivious to details regarding the nature of the signal itself, such as the time between two consecutive R peaks, and based on this, the RR interval variability (also termed Heart Rate Variability, HRV), which has been proven to provide additional information and insight on the nature of the behavior of the organ that is providing the raw signal. Common applications for HRV include the diagnosis of common conditions or diseases such as stress detection, sleep quality analysis, and metrics based on RR time series have been extensively used for the prediction of cardiovascular notable events. All taken into consideration, a set of features is likely to be incomplete if it does not consider features from these two different natures - statistical and dynamic.

### 3.3.1. BPM

Beats per minute (BPM) is a measure of the heart rate, calculated as the average number of beats per minute. It is an essential parameter in evaluating heart function and potential abnormalities. BPM is determined by taking the average of the intervals between each beat (segment) of the signal as:

$$BPM = \frac{60(t_N - t_1)}{N} \tag{8}$$

where $t_N$ refers to the time of the final heartbeat in the considered interval, $t_1$ refers to the time of the first heartbeat in the interval, and $N$ is the total number of heartbeats.

### 3.3.2. IBI

The Interbeat Interval (IBI) is the average time interval between successive heartbeats. It reflects the time variation between heartbeats and provides insight into the autonomic nervous system's regulation of the heart. The IBI is computed as:

$$IBI_{avg} = \frac{6000}{HR_{avg}} [ms] \tag{9}$$

with

$$\overline{HR} = \frac{1}{N}\sum_{i=1}^{N} HR_i \tag{10}$$

where $HR_{avg}$ is the average heart rate, and $HR_i$ represents the heart rate at each individual point, $i$. The average heart rate (HR) [54] is computed using a particle filter method.

### 3.3.3. SDNN

The Standard Deviation of NN intervals (SDNN) is a measure of heart rate variability, representing the overall variability of $RR$ intervals. It is a widely used parameter to assess the autonomic nervous system's influence on heart rate and provides information about the total variability of the heart rate as:

$$SDNN = \sqrt{\frac{1}{N}\sum_{i=1}^{N}(RR_i - \overline{RR})} \tag{11}$$

with

$$\overline{RR} = \frac{1}{N}\sum_{i=1}^{N} RR_i \tag{12}$$

where $N$ is the total number of all $RR$ intervals, and $RR_i$ represents the interval between successive heartbeats at each individual point, $i$, while $\overline{RR}$ is the average interval between heartbeats over $N$ total points.

### 3.3.4. SDSD

The Standard Deviation of Successive Differences (SDSD) is a measure of the variability of consecutive differences between neighboring heartbeat intervals. It is an essential parameter to assess short-term fluctuations in heart rate, which can provide information about the parasympathetic nervous system's influence on the heart as:

$$SDSD = \sqrt{\frac{1}{N-1}\sum_{i=1}^{N-1}\left(|RR_i - RR_{i+1}| - \overline{RRdif}\right)^2} \tag{13}$$

where $RRdif$ is described as following equation

$$\overline{RRdif} = \sum_{i=1}^{N-1}(|RR_i - RR_{i+1}|) \tag{14}$$

### 3.3.5. RMSSD

The Root Mean Square of Successive Differences (RMSSD) is a measure of the short-term variability of RR intervals. It is commonly used to assess the parasympathetic nervous system's influence on heart rate variability as:

$$RMSSD = \sqrt{\frac{1}{N-2}\sum_{i=0}^{N-2}(RR_i - RR_{i+1})^2}$$

(15)

### 3.3.6. PNN50, PNN20

PNN50 and PNN20 are the percentages of time gaps between consecutive heartbeats that are greater than 50 milliseconds and 20 milliseconds, respectively. They are useful for evaluating short-term variations in heart rate and are associated with the parasympathetic nervous system's activity as:

$$NN50 = \sum_{i=1}^{N}(|RR_{i+1} - RR_i| > 50ms)$$

(16)

$$NN20 = \sum_{i=1}^{N}(|RR_{i+1} - RR_i| > 20ms)$$

(17)

where

$$pNN50 = \frac{NN50}{N}.100$$

(18)

$$pNN20 = \frac{NN20}{N}.100$$

(19)

### 3.3.7. HR_MAD

The Median Absolute Deviation (HR_MAD) is a measure of the dispersion of RR intervals, which quantifies the variability of the heart rate. It is more robust to outliers than the standard deviation and can be used to assess the overall variability of the heart rate as:

$$HR\_MAD = \frac{1}{N}\sum_{i=1}^{N}(|RR_i - Median|)$$

(20)

### 3.3.8. SR1, SR2, S and Ratio

SD1 represents the standard deviation measured perpendicular to the identity line [55], while SD2 is the standard deviation along the identity line. S denotes the area of the ellipse that SD1 and SD2 depict, and Ratio is SD1/SD2 as:

$$x = \{x_1, x_2, \ldots, x_n\} = \{RR_1, RR_2, \ldots, RR_n\}$$
$$y = \{y_1, y_2, \ldots, y_n\} = \{RR_2, RR_3, \ldots, RR_{n+1}\}$$

$$SD1 = \sqrt{var(d_1)} \quad ; \quad SD2 = \sqrt{var(d_2)}$$

(21)

$$Ratio = \frac{SD1}{SD2}$$

(22)

where $var(d_x)$ is the variance of $d$

$$d_1 = \frac{x-y}{\sqrt{2}} \quad , \quad d_2 = \frac{x-y}{\sqrt{2}} ,$$

(23)

### 3.3.9. Breathing Rate

Breathing Rate is another essential feature that can be extracted from the ECG signal, providing valuable information about respiratory activity. Breathing rate, often measured in breaths per minute (BPM), is the number of breaths taken within a one-minute interval. It is an important physiological parameter that can help assess overall health, fitness levels, and stress response.

### 3.4. MLP boosted by an improved DE algorithm

This section elaborates on our proposed MLP training algorithm based on an improved DE algorithm for ECG signal classification. The input of MLP is the features extracted, and the output is fed into the next step, local search. The main components of our proposed MLP are explained below.

### 3.4.1. Representation

MLP's weights and biases are stored in a one-dimensional array, shown in Figure 5. The array's length equals the combined weights and biases of the MLP.

### 3.4.2. Objective Function

To achieve our goals, we employ a classification error-based objective function specified as:

$$E = \frac{100}{P} \sum_{p=1}^{P} \xi(x_p)$$

(24)

with

$$\xi(x_p) = \begin{cases} 1 & if \ o_p \neq d_p \\ 0 & otherwise \end{cases}$$

where $x_p$, the $p$-th of $P$ test samples, is input and $d_p$ and $o_p$ are the actual and expected values, respectively.

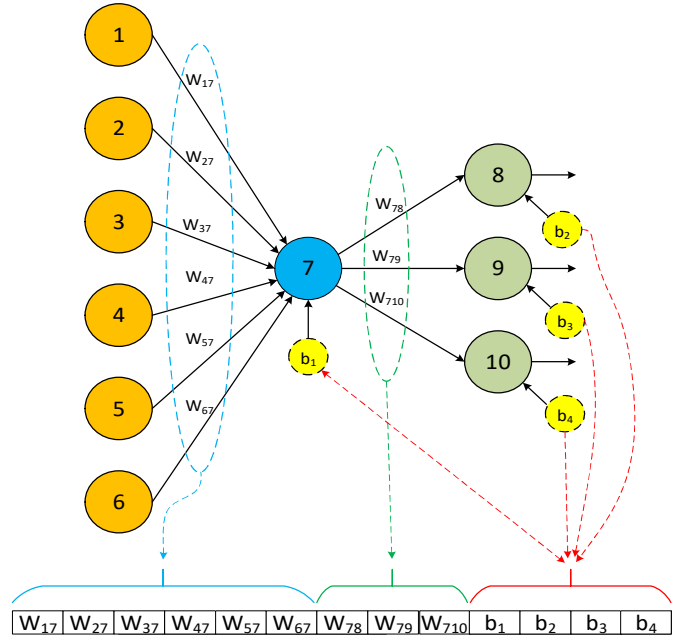

Figure 5: Candidate solution representation in the CODEL algorithm. Top: the network; Bottom: the candidate solutions resultant structure.

### 3.4.3. Cluster-Based Strategy

A clustering method is used to create the regions in our boosted DE algorithm. We accomplished this using the well-known clustering method known as the $k$-means algorithm [56]. A random number between 2 and $\sqrt{N_P}$ is used to determine how many clusters there are. The center of a cluster may be viewed as a crossover of multiple parents as it encompasses all potential solutions within that particular cluster. For the updating process, the general population-based algorithm (GPBA) suggested in [57] is used as follows:

- *Selection:* Candidate solutions should be chosen at random from the population at large. This is equivalent to the $k$-means algorithm's starting point.

- *Generation:* Produce m candidate solutions (set $A$). Using the $k$-means algorithm, CODEL generates new candidate solutions, and each cluster center identifies a new candidate solution.

- *Substitution:* Select m possible solutions (set $A$) for substitution from the present population. CODEL selects $m$ possible solutions at random.

- *Update:* From the collection of possible combinations $A \cup B$, select the $m$ best candidate solutions as $B$. The resulting population is calculated as $(P - B) \cup B$.

Each iteration of CODEL does not apply a clustering method. In other words, clustering algorithms are executed regularly based on a clustering period parameter, similar to the technique provided in [58,59].

### 3.4.4. Opposition-based approach

In order to enhance development, CODEL supports the updating process and initialization through opposition-based learning (OBL) [60]. The definition of $x$ opposition number is:

$$\overline{x_i} = a_i + b_i - x_i \tag{25}$$

where $a$ and $b$ represent the minimum and maximum limits, respectively, and $i$ represents the $i$-th dimension of $x$.

Instead of using regular OBL, this paper utilizes the Quasi-OBL (QOBL) technique as it is more likely to converge on the global solution [61]. Quasi-opposition numbers are defined as:

$$\check{x}_i = rand \left[ \frac{a_i + b_i}{2}, a_i + b_i - x_i \right] \tag{26}$$

where $rand[m, n]$ generates a uniformly random number between $m$ and $n$.

During the initialization process, CODEL generates a random population $Pop$, followed by a corresponding population $OPop$ using the QOBL technique. $Pop$ and $OPop$ are combined to determine the best candidate solutions. The updating step utilizes a similar mechanism.

A new population $OPop$ is created utilizing the region-based technique, the QOBL strategy, and a probability $J_r$ (jumping rate) between 0 and 0.4. The best candidate solutions from the union of $Pop$ and $OPop$ then make up the new population. Algorithm 1 shows the QOBL approach in action using pseudo-code.

---

### Algorithm 1: Pseudocode for QOBL

Input: Problem Parameters
Output: Updated Weights
1. **Procedure** $QOBL$
2.    Initialize   $D$ // dimensionality
3.        $N_p$ // population size
4.        $Pop$ // initial population
5.        $L$ // lower bound
6.        $U$ // upper bound
7. **For** $i$ from 0 to $N_p$ **do**
8.    **For** $i$ from 0 to $D$ **do**
9.        $OPop(i,j) = rand \left[ \frac{L(i,j)+U(i,j)}{2}, (L_{i,j} + U_{i,j} - Pop(i,j)) \right]$
10.    **End for**
11. **End for**
12. Evaluate the value of the objective function for each candidate solution based on Eq. (24)
13. $Pop \leftarrow$ Select $N_p$ best candidate solutions from set $\{Pop, OPop\}$ as initial population
14. **End Procedure**

---

*3.4.5. General structure of the proposed DE trainer*

This paper proposes a DE-based MLP training for ECG signal classification. To this end, we also enhance DE based on a local refinement, a region-based strategy (RBS), and quasi-opposition-based learning (QOBL). RBS tries to improve the exploitation of the DE algorithm, while QOBL comes to enhance the exploration of the proposed algorithm. Local refinement is also used to improve DE trainer based on a gradient-based algorithm in the last step. For local refinement step, we have used 6 algorithms, including Resilient Backpropagation (RP), One-Step Secant (OSS), Gradient Descent (GD), Gradient Descent with Momentum (GDM), Gradient Descent with Adaptive learning rate (GDA) and Conjugate Gradient Backpropagation with Polak-Ribiére Update algorithm (CG-PR). Therefore, this paper introduces six DE-based trainer, named CODEL-RP, CODEL-OSS, CODEL-GD, CODEL-GDM, CODEL-GDA, and CODEL-CG-PR. The proposed algorithm is depicted in detail in Algorithm 2.

---

**Algorithm 2: Pseudocode for CODEL**

Input: Problem Parameters, ECG signals
Output: Updated Weights

   **1.**   **Begin**
   **2.**   Initialize $Npop$ // *population size*
               $NFE_{max}$ // *maximum number of function evaluations*
               $J_r$ // *jumping rate*
               $C_p$ // *clustering period*
               $NFE \leftarrow 0$ // *current number of function evaluations*
               $iter \leftarrow 1$ // *current iteration*
   *3.*   Generate $Pop$ using uniformly distributed random numbers
   *4.*   Compute $Pop$ using Eq. (24) // *objective function of each candidate solution*
   *5.*   Initialize $QOBL\ (D, Np, Pop, L, U)$ for generation a new $OPop$
   *6.*   $NFE \leftarrow NFE + 2 \times N_{pop}$
   *7.*   For each candidate solution
   **8.**      Apply the mutation operator
   **9.**      Apply the crossover operator
   **10.**     Calculate the objective function by utilizing Equation (24).
   **11.**     Apply the selection operator
   **12.**   End for
   *13.*   $NFE \leftarrow NFE + N_{pop}$
   **14.**   If $rem(iter, C_p == 0)$
   **15.**     $k = rand(2, \sqrt{N_p})$
   **16.**     Apply $k$-means clustering on the current population
   **17.**     Select cluster centres as a new set called $A$.
   **18.**     Select $k$ candidate solutions randomly from the current population as $B$.
   **19.**     From $A \cup B$, select best $k$ candidate solutions as $\overline{B}$.
   **20.**     Select new population as $(P - B) \cup \overline{B}$.
   **21.**   End if
   **22.**   If $rand(0,1) < J_r$
   *23.*     Initialize $QOBL\ (D, Np, Pop, L, U)$
   *24.*     $NFE \leftarrow NFE + N_{pop}$
   **25.**   End if
   **26.**   $iter \leftarrow iter + 1$
   **27.**   If $NFE < NFE_{max}$
   **28.**     Do Step 7.
   *29.*   End If
   *30.*   **End**

---

*3.5. Local Search*

Local search is the final step in our proposed algorithm. To this end, we have used six gradient-based algorithms including Resilient Backpropagation (RP), One-Step Secant (OSS), Gradient Descent (GD), Gradient Descent with Momentum (GDM), Gradient Descent with Adaptive learning rate (GDA) and Conjugate Gradient Backpropagation with Polak-Ribiére Update algorithm (CG-PR). In other words, the weights obtained by CODEL are fed as the initialization step to the gradient-based algorithms. Algorithm 3 explains the general structure of the local search.

---

**Algorithm 3: Pseudocode for Local Search**

---

Inputs: weights achieved by CODEL
Output: updated weights

1.   **Begin**
2.   Initialize *sel ← select one approach (1:RP, 2:OSS, 3:GD, 4:GDM, 5:GDA, 6:CG-PR)*
         *The optimal candidate solution within the current population, denoted by ω,*
         *is assigned as the initial weights for performing a local search.*
3.   **Switch** (sel):
4.      **Case** 1:
5.         *// CODEL-RP algorithm*
6.         Run RP using the initialized $\omega$ (Refer to Section 3-5-1)
7.         Break
8.      **Case** 2:
9.         *// CODEL-OSS algorithm*
10.        Run OSS using the initialized $\omega$ (Refer to Section 3-5-2)
11.        Break
12.     **Case** 3:
13.        *// CODEL-GD algorithm*
14.        Run GD using the initialized $\omega$ (Refer to Section 3-5-3)
15.        Break
16.     **Case** 4:
17.        *// CODEL-GDM algorithm*
18.        Run GDM using the initialized $\omega$ (Refer to Section 3-5-4)
19.        Break
20.     **Case** 5:
21.        *// CODEL-GDA algorithm*
22.        Run GDA using the initialized $\omega$ (Refer to Section 3-5-5)
23.        Break
24.     **Case** 6:
25.        *// CODEL-CG-PR algorithm*
26.        Run CG-PR using the initialized $\omega$ (Refer to Section 3-5-6)
27.        Break
28.     **End Switch**
29.  **End**

---

*3.5.1. Resilient Backpropagation (RP)*

RP[62] is a heuristic learning technique that increases the convergence speed by using the weight update's error function's only sign, not its magnitude, of the derivative [63]. The RP algorithm is a neural network training algorithm that functions similarly to the traditional backpropagation algorithm. The updating scheme is defined as:

$$\Delta_{ij}(t) = \begin{cases} \eta^+ . \Delta_{ij}(t-1), & \frac{\partial L}{\partial \omega_{ij}}(t) . \frac{\partial L}{\partial \omega_{ij}}(t-1) > 0 \\ \eta^- . \Delta_{ij}(t-1), & \frac{\partial L}{\partial \omega_{ij}}(t) . \frac{\partial L}{\partial \omega_{ij}}(t-1) < 0 \\ \Delta_{ij}(t-1), & else \end{cases} \quad (27)$$

with

$$\omega_{t+1} = \begin{cases} \omega_t + \Delta_{ij}(t), & \frac{\partial L}{\partial \omega_{ij}}(t) < 0 \\ \omega_t - \Delta_{ij}(t), & \frac{\partial L}{\partial \omega_{ij}}(t) > 0 \\ 0, & else \end{cases} \quad (28)$$

In which $0 < \eta^- < 1 < \eta^+$, where $L$ is error function of the model, and $\omega_{ij}$ denotes the weight from $i$th neuron to $j$th neuron.

### 3.5.2. One-Step Secant (OSS)

A compromise between quasi-Newton algorithms and conjugate gradient algorithms is one step secant backpropagation (OSS) [64]. OSS backpropagation assumes that the preceding Hessian matrix was an identity matrix rather than storing the entire Hessian matrix [65]. Therefore, inverse matrix calculation is not required. The algorithm's storage and calculation costs are decreased by this feature. We have chosen to use the OSS algorithm for optimization purposes. Although scaled conjugate gradient and quasi-Newton algorithms are known to be more accurate, [66], we opted for OSS due to its balance between computational efficiency and accuracy. Additionally, the OSS algorithm is a suitable choice for our problem since it offers a compromise between quasi-Newton and conjugate gradient algorithms, making it a versatile and effective method for various optimization scenarios. Furthermore, its one-step secant update allows for faster convergence, which is beneficial in the context of ECG signal classification

Backpropagation is used to calculate derivatives of performance with respect to the weight and bias vectors $M$. Each vector $M_i$ is adjusted as

$$M = M + a(dM) \quad (29)$$

where dM is the search direction, and a is the parameter chosen to minimize performance along the search direction. The line search function is used to locate the minimum point. The initial search direction is the negative of the performance gradient. In succeeding iterations, the search direction is computed from the new gradient and the previous steps and gradients as

$$dM = -gM + Ac(M_{step}) + Bc(dgM) \quad (30)$$

Here, the gradient is denoted by $gM$, the weights' change is denoted by $M_{step}$, the gradient's change from the previous iteration is denoted by $dgM$, and the combinational scalar products of $gM$, $M_{step}$, and $dgM$ are $Ac$ and $Bc$.

### 3.5.3. Gradient Descent (GD)

The gradient descent backpropagation training technique (GD) is based on minimizing the mean square error between the output of the network and the desired output [67]. A network is considered to have converged and trained when its error reaches a predefined threshold level. Synaptic weights and biases are updated by the GD algorithm, moving along the negative gradient of the error function. The performance function, $Q$, and the weight and bias vectors, $M$, are both calculated using backpropagation. According to the gradient descent as shown in Eq (31), each weight and bias element in vector M is updated as

$$dM = a \times \frac{dQ}{d\chi} \tag{31}$$

where $a$ represents the rate of learning. The weights and biases will change depending on the learning rate, and $\chi$ represents the weight and bias variables in the neural network. The term $dQ / d\chi$ refers to the gradient of the performance function Q with respect to these variables (weights and biases). The weights and biases will change depending on the learning rate, which is compounded by the gradient's inverse. The step is leading to an algorithm becoming unstable increases with the learning rate. Conversely, the longer it takes for the algorithm to converge, the lower the learning rate.

### 3.5.4. Gradient Descent with Momentum (GDM)

This approach for batch steepest descent training frequently offers faster convergence. Momentum enables a network to react to recent trends in the error surface in addition to the local gradient [68]. Momentum enables the network to disregard minor features on the error surface by acting as a low-pass filter. A network could become caught in a shallow local minimum if it lacks impetus. A network can pass through such a minimum if it gains momentum [69].

Two training parameters, the learning rate ($a$) and the momentum constant ($y$), are required by the Gradient descent with momentum(GDM) method. The performance function $Q$'s derivatives with respect to the weight and bias vectors $M$ are computed using backpropagation. According to gradient descent with momentum as described in Eq (32), each vector $M_i$ is changed as

$$dM = y \times \left(dM_{\text{pervious}}\right) + a \times (1 - y) \times \frac{dQ}{dM} \tag{32}$$

where previous $dM_{pervious}$ denotes the prior bias or weight adjustment.

### 3.5.5. Gradient Descent with Adaptive learning rate (GDA)

The learning rate parameter in the common steepest descent (GD) backpropagation algorithm is maintained throughout training. The correct setting of the learning rate parameter, however, has a significant impact on the algorithm's performance. To maintain the largest learning step size while maintaining learning stability, the GDA algorithm was created to enable the learning rate parameter to be adjustable. The ideal value of the learning rate parameter in the GDA method varies with the gradient's path along the error surface.

### 3.5.6. Conjugate gradient backpropagation with Polak-Ribiére updates (CG-PR)

The constant used to determine the new search direction for the Polak-Ribiére update is calculated as the inner product of the prior gradient change and the present gradient divided by the norm squared of the prior gradient [70]. The search direction ($p$) for each iteration of the CG-PR method is determined as

$$p_k = -g_k + \beta_k p_{k-1} \tag{33}$$

where the equation is used to calculate the constant $\boldsymbol{\beta_k}$ for the Polak-Ribiére as

$$\beta_k = \frac{\Delta g_{k-1}^T g_k}{g_{k-1}^T g_{k-1}} \tag{34}$$

where $\beta_k$, is chosen as the vector of magnitude with respect to the error in the previous iteration, $g_k$ is the gradient vector of the error function at the $k$ th epoch, and $g_{k-1}^T g_{k-1}$ is the norm square of the pervious gradient.

## 4. Experimental Results

This section presents the evaluation of our proposed model for ECG signal classification, focusing on its performance and robustness. We utilized the PTB-XL dataset, which comprises 21,799 clinical 12-lead ECGs. After preprocessing, we extracted 13 features, including BPM, IBI, SDNN, SDSD, RMSSD, PNN20, PNN50, HR_MAD, SD1, SD2, S, SD1/SD2, and breathing rate. Subsequently, we employed the MLP boosted by differential evolution and local search to establish the relationship between the input features and the classes. For the local search, we utilized six local search algorithms, which we refer to as CODEL-RP, CODEL-OSS, CODEL-GDM, CODEL-GDA, CODEL-GD, and CODEL-CG-PR in our experimental results.

All population-based metaheuristic algorithms (PBMHs) employed the same standard setting of 25,000 function evaluations, comparable to the number of iterations for conventional algorithms. Each PBMH had a population size of 50. In the case of CODEL, we assumed a crossover probability of 0.9, a scaling factor of 0.5, and a jumping rate of 0.3. Additionally, we set the clustering period and regularization parameters to 10 and 0.1, respectively.

### *4.1. Evaluation Criteria*

We used a set of assessment measures, including accuracy, sensitivity, specificity, precision, F-score, and G-mean, to evaluate the effectiveness of our suggested strategy for classifying ECG signals. These metrics provide a thorough assessment of the classifier's capacity to properly identify both positive and negative examples, making them crucial for comprehending the usefulness and dependability of our classifier in the context of ECG analysis. These metrics are computed using the confusion matrix, which includes true positives (TP), true negatives (TN), false positives (FP), and false negatives (FN) [71].

- True Positives (TP): The number of instances that were correctly identified as positive by the classifier.
- True Negatives (TN): The number of instances that were correctly identified as negative by the classifier.
- False Positives (FP): The number of instances that were incorrectly identified as positive by the classifier.
- False Negatives (FN): The number of instances that were incorrectly identified as negative by the classifier.

The metrics are defined as follows:

1. Accuracy: The proportion of correctly classified instances among the total instances in the dataset. Accuracy measures the overall effectiveness of the classifier.

$$Accuracy = \frac{TP + TN}{TP + TN + FP + FN} \tag{35}$$

2. Sensitivity: The proportion of true positive instances among the actual positive instances. Sensitivity measures the classifier's ability to correctly identify positive cases.

$$Sensitivity = \frac{TP}{TP + FN} \tag{36}$$

3. Specificity: The proportion of true negative instances among the actual negative instances. Specificity measures the classifier's ability to correctly identify negative cases.

$$Specificity = \frac{TN}{TN + FP} \tag{37}$$

4. Precision: The proportion of true positive instances among the predicted positive instances. Precision measures the classifier's ability to accurately classify positive cases as positive.

$$Precisoin = \frac{TP}{TP + FP} \tag{38}$$

5. F-score: The harmonic mean of precision and sensitivity. F-score provides a balanced measure of the classifier's performance, considering both false positives and false negatives.

$$F - Score = \frac{2TP}{2TP + FP + FN} \tag{39}$$

6.  G-mean: The geometric mean of sensitivity and specificity. G-mean is a valuable metric for evaluating classifiers, as it takes into account both the classifier's ability to correctly identify positive cases and its ability to correctly identify negative cases. By considering these two aspects of classification performance, G-mean offers a more comprehensive understanding of the classifier's effectiveness in differentiating between classes.

$$G - Mean = \sqrt{Recall \ \times Specificity}$$ (40)

In addition to the previously mentioned evaluation criteria, we also utilized the win/tie/loss (w/t/l) metric, which indicates whether the local-search-based algorithm in the corresponding row outperforms, performs equally to, or underperforms compared to the original algorithm.

Another evaluation criterion introduced in this paper is the Error Enhancement for each algorithm. It is defined as follows:

$$Error_{alg}(\beta) = 100 - \beta_{alg}$$ (41)

Here, $alg$ represents one of the six base algorithms or six CODEL algorithms, and $\beta$ signifies a measure from one of the selected evaluation criteria for assessments.

$$EE = \frac{Error_{base\_alg}(\beta) - Error_{CODEL\_alg}(\beta)}{Error_{base\_alg}(\beta)} * 100$$ (42)

In this equation, $EE$ represents the Error Enhancement, which provides a measure of the improvement in error rates achieved by the CODEL algorithms compared to the base algorithms. By including this evaluation criterion, we aim to understand better the performance improvements offered by the proposed local search-based methods in the context of ECG signal classification.

The importance of these evaluation criteria in ECG classification and this research lies in their ability to provide a thorough understanding of the classifier's performance across multiple aspects. By considering these metrics, we can identify the strengths and weaknesses of our proposed method and make informed decisions about potential improvements and comparisons with other classification techniques.

### 4.2. Results and Discussion

In this section, an extensive series of numerical experiments are presented to evaluate the efficiency of the CODEL method. Specifically, the evaluation process employs Mean, standard deviation (Std), minimum, maximum, median, rank, and W/T/L criteria. The accuracy, sensitivity, specificity, precision, F-score, and G-mean of six algorithms are compared, including resilient backpropagation (RP) [72], one-step secant backpropagation (OSS) [73], gradient descent with momentum backpropagation (GDM) [74], gradient descent with adaptive learning rate backpropagation (GDA) [75], gradient descent backpropagation training technique (GA) [76], and conjugate gradient backpropagation with Polak-Ribiere updates (CG-PR) [77,78] with their respective improved versions boosted by the CODEL method.

Additionally, the k-fold cross-validation method is employed to assess the effectiveness of the proposed algorithms. In this approach, the dataset is divided into k equal parts. The algorithm is executed ten times, each time selecting a different subset of the dataset as the test set and using the remaining k-1 subsets as the training set. In this study, k is assumed to be equal to 10 (10cv).

In the first, six conventional algorithms, including RP, OSS, GDM, GDA, GA, and CG-PR, are compared with their CODEL-boosted counterparts using six well-established criteria containing Accuracy, Sensitivity, Specificity, Precision, F-score, and G-mean. This comparison aims to demonstrate the advantages of incorporating the CODEL method into these algorithms to improve their ECG signal classification performance.

In the second part of the analysis, we concentrate on examining the Error Enhancement (EE) of the CODEL-boosted algorithms in comparison to their conventional counterparts. The EE metric quantifies the improvement in performance attained by integrating the CODEL method into the original algorithms. By assessing the EE, we aim to acquire a more profound understanding of the effectiveness of the CODEL method in augmenting the performance of ECG signal classification algorithms. This comprehensive analysis will help elucidate the potential benefits of incorporating the CODEL method and provide valuable insights for future research in this area.

In the final part of the analysis, the ranking of each algorithm, enhanced by the CODEL method, is presented. This ranking is based on the performance metrics discussed earlier, including accuracy, sensitivity, specificity, precision, F-score, and G-mean. The algorithms are ranked in descending order of their overall performance, with the top-ranked algorithm exhibiting the most superior performance in ECG signal classification.

By showcasing the ranking of the CODEL-boosted algorithms, we emphasize the enhancements achieved through the integration of the CODEL method into these algorithms. This comparison underlines the CODEL method's effectiveness in augmenting the performance of ECG signal classification algorithms, providing a clear direction for future research endeavors in this area.

Table I presents the results of a 10-fold cross-validation (10CV) experiment on the accuracy measure for all algorithms. The results indicate that the CODEL-boosted algorithms generally outperform their base counterparts. The most significant improvement in mean accuracy is observed for the CODEL-OSS algorithm, which outperforms the original OSS algorithm by approximately 4.48%. CODEL-RP, with a mean accuracy of 71.13%, ranks 6th among all algorithms and shows improvement over the base RP algorithm. CODEL-OSS ranks 3rd with a mean accuracy of 75.36%, demonstrating significant improvement compared to the base OSS algorithm. CODEL-GDM and CODEL-GDA also show notable improvements in accuracy over their base versions, with CODEL-GDM ranking 4th at 72.11% mean accuracy and CODEL-GDA ranking 5th at 71.23% mean accuracy. CODEL-GD, with a mean accuracy of 71.06%, ranks 7th and shows an improvement over the base GD algorithm.

TABLE I
10CV Accuracy measure for all algorithms

| Algorithms | Mean | Std. | Min. | Max. | Med. | Rank | W/T/L |
|---|---|---|---|---|---|---|---|
| **RP** | 70.46 | 1.5 | 68.53 | 73.36 | 70.63 | 9 | W |
| **CODEL-RP** | 71.13 | 1.68 | 68.53 | 74.77 | 70.98 | 6 | |
| **OSS** | 70.88 | 2.45 | 66.9 | 75.7 | 70.36 | 8 | W |
| **CODEL-OSS** | 75.36 | 4.32 | 68.22 | 81.54 | 75.73 | 3 | |
| **GDM** | 68.21 | 2.69 | 64.95 | 74.3 | 67.68 | 12 | W |
| **CODEL-GDM** | 72.11 | 3.06 | 68.46 | 77.8 | 71.45 | 4 | |
| **GDA** | 69.33 | 7.56 | 60.28 | 78.32 | 70.79 | 11 | W |
| **CODEL-GDA** | 71.23 | 3.15 | 66.43 | 76.22 | 70.83 | 5 | |
| **GD** | 69.43 | 1.6 | 67.29 | 72.66 | 69.43 | 10 | W |
| **CODEL-GD** | 71.06 | 1.69 | 68.07 | 73.6 | 71.3 | 7 | |
| **CG-PR** | 79.02 | 2.41 | 74.13 | 83.64 | 78.79 | 2 | W |
| **CODEL-CG-PR** | 79.21 | 2.47 | 74.59 | 82.52 | 79.46 | 1 | |
| **Total W/T/L** | | | | | | | 6/0/0 |

By considering the standard deviation (Std), the most significant values can be seen for GDA (7.56) and CODEL-boosted OSS (4.32), which implies that the accuracy performance of these algorithms might exhibit substantial fluctuations under various conditions. On the other hand, the lowest standard deviation values are found for RP (1.5) and GD (1.6), indicating more consistent performance across the various folds of the cross-validation process. The most significant change in standard deviation between base and CODEL-boosted algorithms is seen in the OSS and CODEL-boosted OSS pair, with an increase of 1.87 units, suggesting that the variability of the CODEL-boosted OSS algorithm is higher compared to the base OSS algorithm. The W/T/L column clearly indicates the performance improvements achieved by the CODEL method. For all the CODEL-boosted algorithms, there are no ties or losses, indicating that the CODEL-boosted algorithms consistently outperform their base versions. The Total W/T/L row highlights that the CODEL-boosted algorithms won in all six comparisons, further emphasizing the CODEL method's effectiveness in enhancing the base algorithms' performance.

Finally, the CODEL-CG-PR algorithm achieves the highest mean accuracy of 79.21% and ranks 1st among all algorithms. This result indicates that the CODEL method is effective in enhancing the performance of the base CG-PR algorithm, which also performs well with a mean accuracy of 79.02% and a rank of 2nd. In summary, Table I demonstrates that the application of the CODEL method to the base algorithms leads to improved classification accuracy across the board, with each CODEL-boosted algorithm winning against their respective base versions.

Table II presents the results of a 10-fold cross-validation (10CV) experiment on the sensitivity measure for all algorithms. In contrast to the results in Table I, the CODEL-boosted algorithms' performance improvement in sensitivity is not as consistent. While some CODEL-boosted algorithms show better sensitivity than their original counterparts, others exhibit a decrease in sensitivity.

TABLE II
10CV Sensitivity measure for all algorithms

| Algorithms | Mean | Std. | Min. | Max. | Med. | Rank | W/T/L |
|---|---|---|---|---|---|---|---|
| **RP** | 83.42 | 1.97 | 80.31 | 86.1 | 82.82 | 9 | W |
| **CODEL-RP** | 83.89 | 1.76 | 81.08 | 87.26 | 84.14 | 5.5 | |
| **OSS** | 83.77 | 2.07 | 81.01 | 87.64 | 83.2 | 7 | L |
| **CODEL-OSS** | 81.03 | 3.4 | 75.29 | 85.33 | 81.43 | 11 | |
| **GDM** | 83.5 | 4.85 | 75.68 | 89.96 | 82.63 | 8 | W |
| **CODEL-GDM** | 87.91 | 4.32 | 81.08 | 94.98 | 88.39 | 1 | |
| **GDA** | 83.89 | 14.87 | 64.86 | 100 | 87.26 | 5.5 | W |
| **CODEL-GDA** | 84.35 | 4.35 | 78.38 | 90.73 | 82.79 | 3 | |
| **GD** | 83.31 | 1.53 | 80.69 | 86.1 | 83.37 | 10 | W |
| **CODEL-GD** | 84.2 | 2.42 | 79.54 | 87.98 | 84.17 | 4 | |
| **CG-PR** | 84.43 | 2.92 | 79.92 | 87.26 | 86.1 | 2 | L |
| **CODEL-CG-PR** | 79.79 | 1.57 | 77.22 | 83.01 | 79.92 | 12 | |
| **Total W/T/L** | | | | | | | 4/0/2 |

Regarding sensitivity, the CODEL-enhanced algorithms show mixed results compared to their base counterparts. CODEL-RP has a mean sensitivity of 83.89% and ranks 5.5, slightly outperforming the base RP algorithm. However, CODEL-OSS ranks 11th with a mean sensitivity of 81.03%, indicating a decline in performance compared to the base OSS algorithm. On the other hand, CODEL-GDM and CODEL-GDA demonstrate considerable improvements in sensitivity over their base versions. CODEL-GDM achieves the highest mean sensitivity of 87.91% and ranks 1st among all algorithms, while CODEL-GDA ranks 3rd with a mean sensitivity of 84.35%. CODEL-GD also shows improvement over the base GD algorithm, with a mean sensitivity of 84.2% and a rank of 4th.

Interestingly, the CODEL-CG-PR algorithm experiences a decline in performance compared to the base CG-PR algorithm, ranking last with a mean sensitivity of 79.79%. This result suggests that the CODEL method may not always enhance the performance of an algorithm in every performance metric.

The standard deviation (Std) values in Table II show the variation in sensitivity for each algorithm over various cross-validation folds. GDA (14.87) and GDM (4.85) have the biggest standard deviation values. In contrast, CODEL-CG-PR (1.57) and GD (1.53) have the lowest standard deviation values, indicating more consistent performance across the many folds of the cross-validation procedure. The W/T/L column provides insight into the performance improvements achieved by the CODEL method concerning Sensitivity. While the CODEL-boosted algorithms demonstrate mixed results, with four wins, two losses, and no ties, some CODEL-boosted algorithms exhibit significant improvements over their base versions. For example, CODEL-GDM outperforms its base counterpart, GDM, in Sensitivity, ranking 1st with a mean of 87.91%. On the other hand, CODEL-CG-PR shows a decline in performance compared to its base version, ranking 12th with a mean Sensitivity of 79.79%. The Total W/T/L row emphasizes that the overall impact of the CODEL method on Sensitivity is not as clear-cut as it was for Accuracy in Table I, with some algorithms benefiting from the enhancement while others experiencing a decline in performance.

In summary, the results in Table II provide a more nuanced understanding of the CODEL method's impact on the performance of optimization algorithms for ECG signal classification. The CODEL-boosted algorithms exhibit varying performance in terms of sensitivity, indicating the importance of considering multiple evaluation metrics when assessing algorithm performance.

Table III presents the results of a 10-fold cross-validation (10CV) experiment on the specificity measure for all algorithms. In terms of specificity, most of the CODEL-boosted algorithms show improvements when compared to their base counterparts. CODEL-RP demonstrates a mean specificity of 51.68% and ranks 4th, outperforming the base RP algorithm. Similarly, CODEL-OSS achieves a mean specificity of 66.7% and ranks 3rd, indicating a significant improvement over the base OSS algorithm.

CODEL-GDM and CODEL-GDA also exhibit improvements in specificity, with CODEL-GDM ranking 10th with a mean specificity of 48.03% and CODEL-GDA ranking 5th with a mean specificity of 51.22%. CODEL-GD shows a marginal improvement over the base GD algorithm, with a mean specificity of 51.03% and a rank of 7th. The CODEL-CG-PR algorithm significantly outperforms its base version, achieving the highest mean specificity of 78.32% and ranking 1st among all algorithms. This result highlights the effectiveness of the CODEL method in enhancing the specificity of the CG-PR algorithm. Similar to the results in Table I, the W/T/L criteria in Table III indicate that the CODEL-boosted algorithms win against the original algorithms six times, tie 0 times, and lose 0 times.

TABLE III
10CV Specificity measure for all algorithms

| Algorithms | Mean | Std. | Min. | Max. | Med. | Rank | W/T/L |
|---|---|---|---|---|---|---|---|
| **RP** | 50.68 | 3.37 | 44.12 | 54.71 | 51.62 | 8 | W |
| **CODEL-RP** | 51.68 | 3.64 | 45.88 | 60.59 | 51.62 | 4 | |
| **OSS** | 51.21 | 5.19 | 41.76 | 59.17 | 51.33 | 6 | W |
| **CODEL-OSS** | 66.7 | 13.41 | 42.01 | 84.12 | 67.94 | 3 | |
| **GDM** | 44.89 | 12.12 | 30.77 | 66.47 | 43.65 | 12 | W |
| **CODEL-GDM** | 48.03 | 13.21 | 32.94 | 69.23 | 43.53 | 10 | |
| **GDA** | 47.11 | 41.26 | 40.51 | 97.06 | 45.56 | 11 | W |
| **CODEL-GDA** | 51.22 | 12.9 | 29.41 | 66.47 | 50.88 | 5 | |
| **GD** | 48.26 | 3.77 | 40.59 | 54.44 | 47.65 | 9 | W |
| **CODEL-GD** | 51.03 | 3.28 | 45.88 | 56.47 | 51.47 | 7 | |
| **CG-PR** | 70.78 | 5.26 | 63.53 | 79.41 | 71.39 | 2 | W |
| **CODEL-CG-PR** | 78.32 | 5.56 | 66.47 | 85.88 | 79.65 | 1 | |
| **Total W/T/L** | | | | | | | 6/0/0 |

In Table III, the standard deviation (Std) values for Specificity exhibit a wide range of variation across different algorithms. The highest standard deviations are observed for GDA (41.26) and CODEL-OSS (13.41). In contrast, algorithms such as RP (3.37) and CODEL-GD (3.28) exhibit lower standard deviations, indicating more consistent performance in the Specificity metric.

The W/T/L column in Table III provides insights into the performance improvements achieved by the CODEL method for Specificity. All CODEL-boosted algorithms register wins against their respective base algorithms, with no ties or losses, emphasizing the effectiveness of the CODEL method in enhancing the performance of the base algorithms for Specificity.

The results in Table III demonstrate the effectiveness of the CODEL method in improving the performance of various optimization algorithms for ECG signal classification, particularly with respect to specificity. The CODEL-boosted algorithms consistently outperform their original counterparts in this metric. The benefits of incorporating the CODEL approach include enhanced algorithm performance, more consistent results across various conditions, and the potential for improved clinical decision-making in ECG classification studies.

Table IV highlights the impact of the CODEL method on the precision of different algorithms. Precision is an important metric for evaluating classification algorithms, as it measures the proportion of true positive instances among those predicted as positive by the classifier. High precision is crucial in applications where the cost of false positives is high, such as medical diagnosis or fraud detection.

TABLE IV
10CV Precision measure for all algorithms

| Algorithms | Mean | Std. | Min. | Max. | Med. | Rank | W/T/L |
|------------|------|------|------|------|------|------|-------|
| **RP** | 72.09 | 1.33 | 69.75 | 74.09 | 72.59 | 10 | |
| **CODEL-RP** | 72.61 | 1.53 | 70.13 | 76.41 | 72.39 | 6 | W |
| **OSS** | 72.42 | 2.19 | 68.57 | 76.04 | 72.24 | 8 | |
| **CODEL-OSS** | 79.4 | 6.04 | 69.28 | 88.05 | 79.73 | 3 | W |
| **GDM** | 70.18 | 3.99 | 66.09 | 78.24 | 68.63 | 12 | |
| **CODEL-GDM** | 72.53 | 4.6 | 67.89 | 80.6 | 70.65 | 7 | W |
| **GDA** | 76.1 | 15.3 | 60.37 | 97.11 | 71.07 | 4 | |
| **CODEL-GDA** | 72.95 | 4.44 | 66.2 | 78.97 | 72.38 | 5 | L |
| **GD** | 71.09 | 1.49 | 68.44 | 73.99 | 70.99 | 11 | |
| **CODEL-GD** | 72.41 | 1.35 | 69.93 | 74.22 | 72.3 | 9 | W |
| **CG-PR** | 81.59 | 2.65 | 77.21 | 86.43 | 82.3 | 2 | |
| **CODEL-CG-PR** | 84.99 | 3.24 | 78.41 | 89.66 | 85.76 | 1 | W |
| **Total W/T/L** | | | | | | | 5/0/1 |

Analysing Table IV, we can observe that the CODEL method has a positive effect on most algorithms' precision. The improvements in precision achieved by CODEL-boosted algorithms show the effectiveness of the method in minimizing false positives in the predictions. Comparing the mean precision values, we can see that CODEL-CG-PR significantly outperforms the other algorithms with a mean precision of 84.99. This suggests that the CODEL-CG-PR algorithm is more reliable in identifying positive instances correctly.

Furthermore, the standard deviation values in Table IV indicate the variability in precision among different cross-validation folds. The highest standard deviations are observed for GDA (15.3) and CODEL-OSS (6.04), indicating that these algorithms have a higher degree of fluctuation in their Precision performance, possibly due to the sensitivity of their models to the specific data characteristics within each fold. In contrast, the lowest standard deviation values are found for RP (1.33) and CODEL-GD (1.35), suggesting more consistent and stable performance across the cross-validation folds. The W/T/L column in Table IV provides insights into the performance improvements achieved by the CODEL method for Precision. Five out of the six CODEL-boosted algorithms register wins against their respective base algorithms, with only one loss and no ties. This emphasizes the effectiveness of the CODEL method in enhancing the performance of the base algorithms for Precision, further confirming the potential benefits of incorporating the CODEL approach in ECG signal classification studies.

In conclusion, Table IV illustrates the positive impact of the CODEL method on the precision of different algorithms, with CODEL-CG-PR exhibiting the best performance among all the considered classifiers. The results emphasize the significance of the CODEL method in improving classification algorithms, particularly in applications where precision is a critical evaluation metric.

Table V presents the F-Score, a harmonic mean of precision and recall (sensitivity), for each of the considered classification algorithms and their CODEL-boosted counterparts. F-Score is a crucial metric when evaluating classification algorithms, as it provides an even-handed assessment of false positives and negatives. This metric is particularly useful in situations where both types of errors have significant consequences, as it gives equal importance to precision and recall, thus offering a more comprehensive evaluation of the classifier's performance.

From Table V, it is evident that the CODEL method has positively impacted the F-Score of most algorithms. The mean F-Score of the enhanced algorithms is generally higher than their base counterparts. Notably, the CODEL-CG-PR algorithm achieves the second-highest mean F-Score of 82.28, only slightly lower than the base CG-PR algorithm, which has the highest mean F-Score of 82.94. This demonstrates the effectiveness of the CODEL method in balancing precision and recall in classification tasks. In Table V, looking at the standard deviation (Std) values offers insight into the variability of F-Score performance associated with each algorithm. The most significant standard deviation values are observed in CODEL-OSS (2.64) and CG-PR (1.94), indicating that the F-Score performance of these algorithms may show substantial fluctuations under different circumstances. Conversely, the smallest standard deviation values are observed for GD (1.12) and RP (1.17), indicating a more consistent F-Score performance throughout the 10CV process.

TABLE V
10CV F-Score measure for all algorithms

| Algorithms | Mean | Std. | Min. | Max. | Med. | Rank | W/T/L |
|---|---|---|---|---|---|---|---|
| **RP** | 77.32 | 1.17 | 75.5 | 79.64 | 77.37 | 9 | |
| **CODEL-RP** | 77.83 | 1.24 | 76.19 | 80.07 | 77.74 | 7 | W |
| **OSS** | 77.66 | 1.74 | 75.26 | 81.23 | 77.19 | 8 | |
| **CODEL-OSS** | 79.98 | 2.64 | 76.47 | 84.6 | 79.52 | 3 | W |
| **GDM** | 76.05 | 1.47 | 72.83 | 78.85 | 75.98 | 12 | |
| **CODEL-GDM** | 79.25 | 1.39 | 77.08 | 81.97 | 79.64 | 4 | W |
| **GDA** | 76.99 | 1.63 | 75 | 79.66 | 76.73 | 10 | |
| **CODEL-GDA** | 78.02 | 1.6 | 75.49 | 80.75 | 78.11 | 5 | W |
| **GD** | 76.7 | 1.12 | 74.91 | 78.92 | 76.62 | 11 | |
| **CODEL-GD** | 77.84 | 1.4 | 75.05 | 79.93 | 78.11 | 6 | W |
| **CG-PR** | 82.94 | 1.94 | 79.1 | 86.43 | 82.86 | 1 | |
| **CODEL-CG-PR** | 82.28 | 1.87 | 79.16 | 84.73 | 82.33 | 2 | L |
| **Total W/T/L** | | | | | | | 5/0/1 |

The performance improvements achieved by the CODEL method for F-Score are revealed through the W/T/L column in Table V. Among the six comparisons between the base algorithms and their CODEL-boosted counterparts, CODEL-boosted algorithms secured wins in five instances, with only one loss recorded for CODEL-CG-PR against the base CG-PR algorithm. These results further emphasize the effectiveness of the CODEL method in enhancing the performance of the base algorithms in terms of F-Score.

In summary, Table V underscores the positive influence of the CODEL method on the F-Score of different algorithms, with CODEL-CG-PR exhibiting a strong performance among all considered classifiers. These results emphasize the importance of the CODEL method in improving classification algorithms, especially in contexts where balancing precision and recall is critical.

Table VI presents the G-Mean, the geometric mean of sensitivity and specificity, for all classification algorithms and their CODEL-boosted counterparts. G-Mean is a valuable metric in evaluating classification algorithms' performance, as it gives equal consideration to both sensitivity and specificity. This metric is especially useful when assessing the performance of classifiers in contexts, where maintaining an equilibrium between these two aspects is of great importance.

From Table VI, it is evident that the CODEL method has positively impacted the G-Mean of most algorithms, with the enhanced algorithms generally achieving higher mean G-Mean scores than their base counterparts. Specifically, the CODEL-CG-PR algorithm outperforms all others with the highest mean G-Mean of 79 and a rank of 1, demonstrating the effectiveness of the CODEL method in simultaneously optimizing sensitivity and specificity in classification tasks.

TABLE VI
10CV G-Mean measure for all algorithms

| Algorithms | Mean | Std. | Min. | Max. | Med. | Rank | W/T/L |
|---|---|---|---|---|---|---|---|
| **RP** | 64.97 | 2.09 | 61.08 | 68.09 | 65.76 | 7 | W |
| **CODEL-RP** | 65.8 | 2.31 | 61.86 | 71.39 | 65.5 | 4 | |
| **OSS** | 65.41 | 3.49 | 59.02 | 71.26 | 65.33 | 6 | W |
| **CODEL-OSS** | 72.99 | 6.96 | 59.87 | 80.9 | 74.9 | 3 | |
| **GDM** | 60.46 | 6.46 | 52.04 | 72.67 | 59.37 | 11 | W |
| **CODEL-GDM** | 64.17 | 7.23 | 55.47 | 75.98 | 62.29 | 9 | |
| **GDA** | 47.51 | 32.97 | 52.03 | 80.04 | 62.99 | 12 | W |
| **CODEL-GDA** | 64.97 | 7.18 | 51.66 | 74.11 | 64.92 | 7 | |
| **GD** | 63.35 | 2.45 | 58.7 | 67.85 | 63.22 | 10 | W |
| **CODEL-GD** | 65.5 | 2.07 | 61.69 | 68.15 | 65.46 | 5 | |
| **CG-PR** | 77.23 | 2.9 | 71.77 | 82.85 | 77.63 | 2 | W |
| **CODEL-CG-PR** | 79 | 3.01 | 72.89 | 83.05 | 79.49 | 1 | |
| **Total W/T/L** | | | | | | | 6/0/0 |

Examining Table VI, the Std values provide insights into the dispersion of G-Mean performance across different algorithms. Greater standard deviation values indicate a higher level of fluctuation in the performance of the algorithm, while lower values suggest more consistent performance. For instance, CODEL-OSS and GDA show notably higher standard deviation values, suggesting that their performance might vary considerably. On the other hand, algorithms such as GD and RP display lower standard deviation values, which indicates that their G-Mean performance is relatively more stable and consistent throughout the 10CV process.

The W/T/L criteria, which compare the performance of the base algorithms and their CODEL-boosted versions, reveal that the enhanced algorithms outperform their base counterparts in 6 cases. This further supports the effectiveness of the CODEL method in optimizing the G-Mean of classification algorithms.

In summary, Table VI highlights the positive influence of the CODEL method on the G-Mean of different algorithms, with CODEL-CG-PR exhibiting the strongest performance among all considered classifiers. These results emphasize the importance of the CODEL method in improving classification algorithms, especially in contexts where balancing sensitivity and specificity is crucial.

In conclusion, after examining the presented tables (Table I-VI), the boosted CODEL algorithms demonstrate a general trend of improvement across various evaluation metrics. However, the degree of improvement varies among different algorithms and metrics. The most significant enhancements are observed in Sensitivity (CODEL-GDM) and Specificity (CODEL-CG-PR), while the improvements for CODEL-GDA in several metrics seem to be less pronounced. On the other hand, some algorithms exhibit minor deterioration in specific metrics. Overall, the average performance of the boosted CODEL algorithms indicates a positive trend, contributing to their effectiveness in the evaluated tasks.

TABLE VII
Performance Evaluation of CODEL Algorithms Across Metrics

| Algorithms | | ACCURACY | SENSITIVITY | SPECIFICITY | PRECISION | F-SCORE | G-MEAN |
|---|---|---|---|---|---|---|---|
| **CODEL-RP** | Mean | 71.13 | 83.89 | 51.68 | 71.61 | 77.83 | 65.80 |
| | Rank | 5 | 4 | 3 | 6 | 6 | 3 |
| **CODEL-OSS** | Rank | 75.36 | 81.03 | 66.70 | 79.40 | 79.98 | 72.99 |
| | Mean | 2 | 5 | 2 | 2 | 2 | 2 |
| **CODEL-GDM** | Rank | 72.11 | 87.91 | 48.03 | 72.53 | 79.25 | 64.17 |
| | Mean | 3 | 1 | 6 | 4 | 3 | 6 |
| **CODEL-GDA** | Rank | 71.23 | 84.35 | 51.22 | 72.95 | 78.02 | 64.97 |
| | Mean | 4 | 2 | 4 | 3 | 4 | 5 |
| **CODEL-GD** | Rank | 71.06 | 84.20 | 51.03 | 72.41 | 77.84 | 65.50 |
| | Mean | 6 | 3 | 5 | 5 | 5 | 4 |
| **CODEL-CG-PR** | Rank | 79.21 | 79.79 | 78.32 | 84.99 | 82.28 | 79.00 |
| | Mean | 1 | 6 | 1 | 1 | 1 | 1 |

Table VII presents the the performance of each improved algorithm by comparing their performances across different evaluation metrics, including accuracy, sensitivity, specificity, precision, F-score, and G-mean. The mean values of the metrics and the ranks of the algorithms are displayed in each row. The lower the rank, the better the performance of the algorithm for that specific metric.

The results in Table VII demonstrate that the CODEL method leads to significant improvements in various performance metrics across all six algorithms. For instance, the CODEL-CG-PR algorithm ranks first in five out of the six metrics: accuracy, specificity, precision, F-Score, and G-Mean. This demonstrates the CODEL method's robustness and effectiveness in improving the base algorithms' performance.

Table VII unequivocally demonstrates how the CODEL method has improved the majority of the base algorithms in terms of the metrics taken into consideration. Five metrics, including accuracy, specificity, precision, F-Score, and G-Mean, place the CODEL-CG-PR algorithm in the top spot. This demonstrates how the CODEL method can be used to optimize various aspects of classification performance.

It is important to take into account multiple evaluation criteria when evaluating the performance of classification algorithms, as their rankings can differ depending on the metric used. For instance, while CODEL-GDM performs well in sensitivity, it ranks lower in specificity and G-mean. This suggests that even though the algorithm is good at identifying true positives, it might have trouble correctly classifying negatives.

Briefly, Table VII thoroughly evaluates the boosted algorithms using several performance metrics, demonstrating the efficiency of the CODEL in boosting the accuracy of different algorithms. The ranking provides a better understanding of the advantages and disadvantages of each algorithm and highlights the necessity of evaluating classification algorithms based on multiple factors.

Figure 6 demonstrates that the CODEL-CG-PR algorithm outperforms other CODEL-enhanced algorithms with a mean rank of 3.0, indicating its superior performance across various performance metrics for ECG signal classification tasks. This suggests that the CODEL method is effective in enhancing the performance of the CG-PR algorithm. In terms of rankings, the CODEL-OSS algorithm comes in second place with a mean rank of 4.3. Next in the ranking in sequence are CODEL-GDA with a mean rank of 5.0, CODEL-RP with a mean rank of 5.4, CODEL-GDM with a mean rank of 5.8, and finally, CODEL-GA with a mean rank of 6.3. The findings indicate that algorithm rankings may differ based on the selected performance measure, highlighting the significance of considering various performance metrics when evaluating and choosing the best-suited algorithm for a particular task.

As seen in Figure 6, which demonstrates the mean ranks of the six improved algorithms by CODEL, the overall effectiveness of the CODEL method in enhancing the performance of these algorithms in ECG signal classification tasks can be seen. The CODEL-CG-PR algorithm, with the best mean rank, can be considered an excellent choice for ECG signal classification, while the other CODEL-enhanced algorithms may offer suitable alternatives depending on the specific requirements and priorities of a given application.

Figure 7 shows a line chart depicting the mean values of the evaluation metrics for each algorithm. It is evident from the plot that the CODEL-boosted versions generally outperform their non-boosted counterparts in terms of accuracy, sensitivity, specificity, precision, F-score, and G-mean. The most significant improvements can be observed in Sensitivity (CODEL-GDM) and Specificity (CODEL-CG-PR), which demonstrates the ability of the CODEL-boosting technique to balance the true positive and true negative rates effectively. However, the extent of improvement varies across different algorithms and metrics, with CODEL-GDA showing a less pronounced enhancement in several metrics.

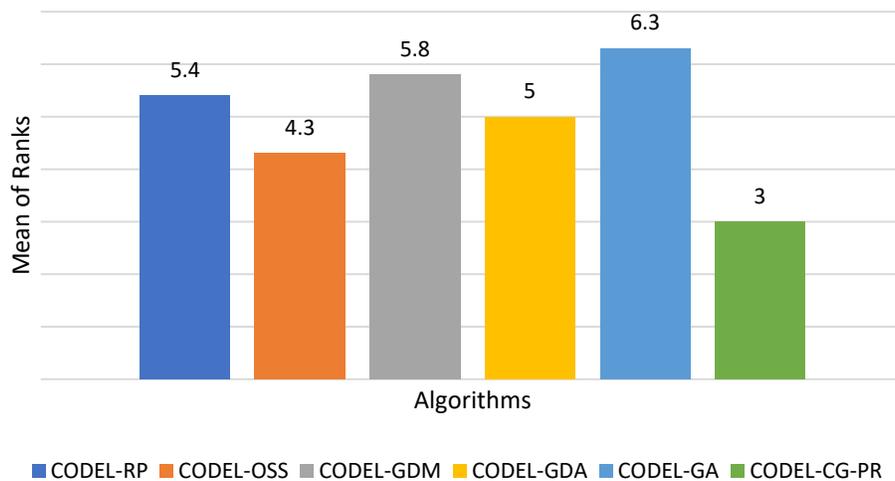

Figure 6: Rank of each algorithms that enhanced by CODEL.

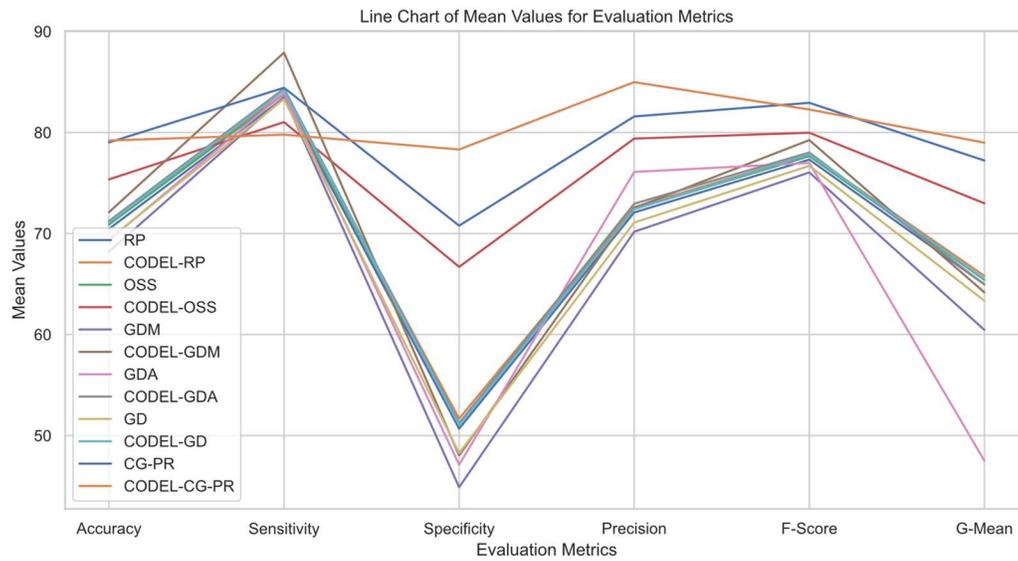

Figure 7: Line Chart Comparing Mean Values of Evaluation Metrics for Original and CODEL-Boosted Algorithms.

Figure 8 presents a grouped bar chart of the mean values for the evaluation metrics, which allows for a clearer visual comparison between the original and CODEL-boosted algorithms. The bar chart further confirms the general trend of improvement across the evaluation metrics, as observed in Figure 7. It is worth noting that, in some cases, the improvement might not be substantial; however, when considering the overall performance in terms of multiple metrics, the CODEL-boosted algorithms demonstrate a consistent and notable enhancement.

Figure 9 displays a box plot of the mean values for the evaluation metrics, providing additional insights into the distribution and variability of the performance measures. The box plot reveals that the CODEL-boosted algorithms not only improve the mean values but also exhibit a reduced spread of the performance metrics, indicating a more stable and consistent performance. This is especially important in real-world applications where consistency is critical to algorithm reliability.

In this research, we have investigated the impact of the CODEL method on different classification algorithms and their performance in the context of ECG data analysis. To evaluate the effectiveness of CODEL-boosted algorithms, we have comprehensively evaluated their performance using multiple metrics, including accuracy, sensitivity, specificity, precision, F-score, G-mean, and EE. The results show acceptable improvements in the CODEL method in different classifiers.

The analysis of the results shows that the CODEL method has positively affected the performance of most of the basic algorithms, leading to more advanced versions that offer better performance in several evaluation criteria. It is worth mentioning that the CODEL-CG-PR algorithm has emerged as the best performer and has been ranked first in accuracy, specificity, precision, F-score, and G-Mean. This shows the effectiveness of the CODEL method in optimizing different aspects of classification in ECG data analysis.

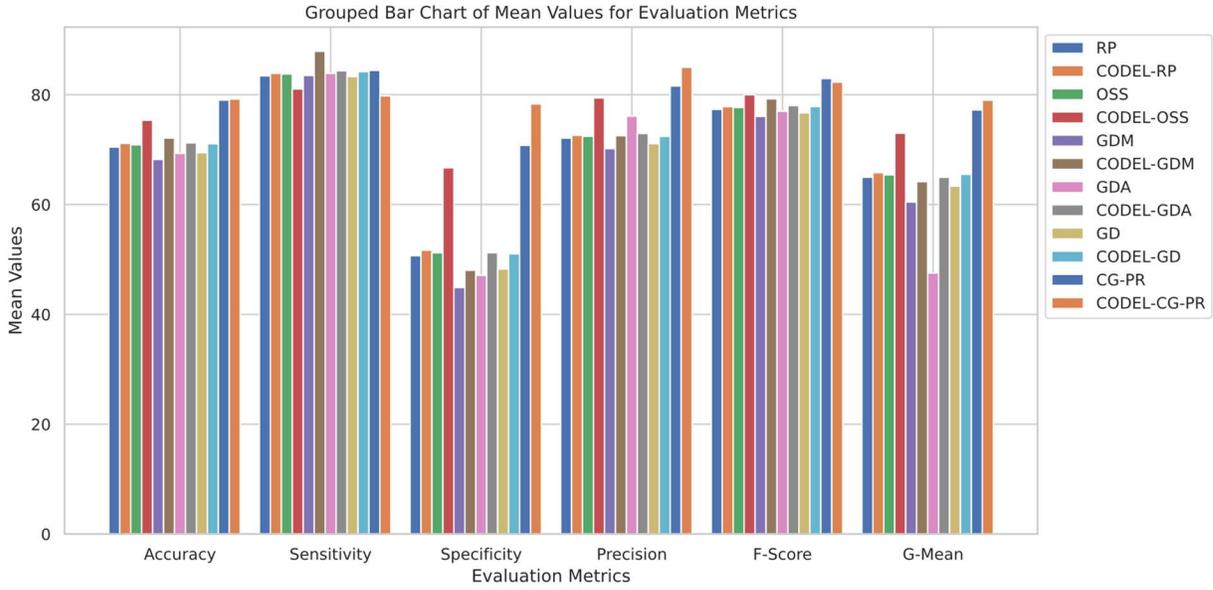

Figure 8: Grouped Bar Chart Illustrating Performance Improvements in Evaluation Metrics for CODEL-Boosted Algorithms.

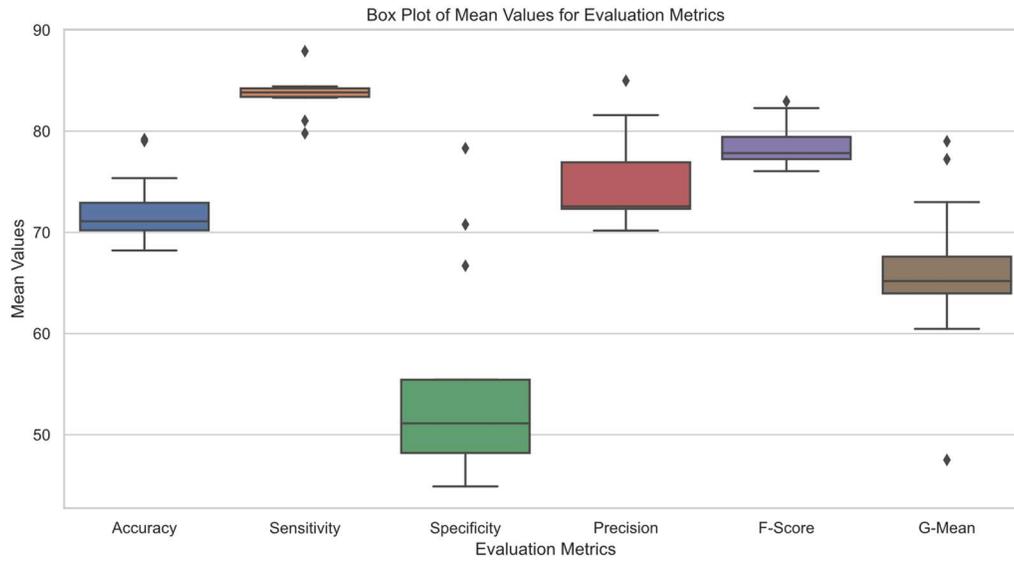

Figure 9: Box Plot Demonstrating the Distribution and Variability of Evaluation Metrics for Original and CODEL-Boosted Algorithms.

The effectiveness of the CODEL method in augmenting the performance of base algorithms is further highlighted by comparing the enhanced algorithms with their original versions. Significant improvements in specificity and G-mean metrics underscore the value of the CODEL method when working with ECG datasets, where striking a balance between sensitivity and specificity is vital for precise diagnosis and treatment planning.

Nonetheless, it's crucial to acknowledge the fluctuating performance of the algorithms across various metrics. The ranks of the algorithms vary depending on the evaluation criteria, emphasizing the need to consider multiple evaluation metrics when assessing classification performance in ECG data analysis. For example, CODEL-GDM ranks first in sensitivity but lower in other metrics, signifying its ability to identify true positives but potential shortcomings in accurately classifying negatives.

In conclusion, the findings reveal the efficacy of the CODEL method in enhancing the performance of different classification algorithms, particularly in ECG data analysis, where balancing sensitivity and specificity is crucial. The thorough evaluation of the improved algorithms highlights the significance of considering multiple evaluation criteria for a well-rounded understanding of their strengths and limitations. The CODEL method has proven to be a valuable tool for optimizing the performance of classification algorithms in ECG data analysis and could be advantageous in situations where accurate detection of cardiac abnormalities and patient-specific treatment planning are essential.

### 4.2.1. *Error Enhancement Analysis of CODEL-Boosted Algorithms*

Table VIII showcases the error enhancement statistics of the CODEL-enhanced algorithms relative to their original base algorithms, as defined by the EE calculation in Equation (42). This table specifically details the percentage improvements in error reduction for various performance metrics. From the data presented in Table VIII, the CODEL-enhanced methods generally exhibit an improvement over their respective base algorithms across multiple metrics, indicating the efficiency of the CODEL approach in optimizing error rates. For example, the CODEL-OSS algorithm registers a substantial enhancement in specificity (31.77%), precision (25.32%), and G-Mean (21.93%), making it rank first in those metrics. Conversely, it's worth noting certain metrics where the enhancement wasn't consistent, as seen in the negative values for CODEL-CG-PR in accuracy (-29.81%) and f-score (-3.87%), CODEL-OSS in sensitivity (-16.86%), and CODEL-GDA in precision (-13.18%). This illustrates that while the CODEL method does provide overall improvements, there might be specific areas where fine-tuning or further adaptations are necessary. Interestingly, CODEL-GDA showcases a remarkable 33.25% improvement in G-Mean, emphasizing its heightened performance in terms of balanced sensitivity and specificity.

The varied rankings across the algorithms for different metrics underline the importance of considering multiple evaluation parameters to gain a comprehensive understanding of an algorithm's enhancement. Table VIII provides invaluable insights into the benefits of the CODEL method in reducing errors, emphasizing the potential it holds for refining classification algorithm performances.

As illustrated in Table VII, the CODEL-CG-PR algorithm emerged as a top performer, securing the highest rankings in most of the metrics. However, when examining Table VIII, the scenario is somewhat different. Despite its superior performance in general metrics in Table VII, the error enhancement statistics in Table VIII show that the CODEL-CG-PR algorithm does not consistently outperform its peers in terms of error reduction percentages. This discrepancy underscores the importance of evaluating algorithms from multiple angles and using diverse metrics. While an algorithm might excel in one aspect, there might be other facets where there's room for improvement. The contrasting results from the two tables highlight the intricacies of algorithmic performance and the need for holistic evaluations.

TABLE VIII
Error enhancement for each improved algorithm. Each row shows that the corresponding algorithm.

| Algorithms | | ACCURACY | SENSITIVITY | SPECIFICITY | PRECISION | F-SCORE | G-MEAN |
|---|---|---|---|---|---|---|---|
| **CODEL-RP vs RP** | Mean | 2.27% | 2.83% | 2.03% | 1.86% | 2.25% | 2.37% |
| | Rank | 5 | 4 | 6 | 5 | 5 | 6 |
| **CODEL-OSS vs OSS** | Rank | 15.41% | -16.86% | 31.77% | 25.32% | 10.38% | 21.93% |
| | Mean | 1 | 2 | 1 | 1 | 2 | 2 |
| **CODEL-GDM vs GDM** | Rank | 12.24% | 26.73% | 5.69% | 7.86% | 13.38% | 9.38% |
| | Mean | 2 | 1 | 4 | 3 | 1 | 3 |
| **CODEL-GDA va GDA** | Rank | 6.20% | 2.86% | 7.77% | -13.18% | 4.48% | 33.25% |
| | Mean | 3 | 3 | 3 | 6 | 4 | 1 |
| **CODEL-GD vs GD** | Rank | 5.33% | 5.33% | 5.34% | 4.56% | 4.89% | 5.87% |
| | Mean | 4 | 2 | 5 | 4 | 3 | 5 |
| **CODEL-CG-PR vs CG-PR** | Rank | 0.90% | -29.81% | 25.85% | 18.45% | -3.87% | 7.78% |
| | Mean | 6 | 5 | 2 | 2 | 6 | 4 |

### 4.2.2. *Comparison of Different Classifiers*

In this experiment, we selected several classification algorithms to validate the performance of our proposed improved algorithms. We compared them with a variety of state-of-the-art and classical classifiers. The evaluated classifiers are: Gaussian Naive Bayes (GNB), Complement Naive Bayes (CNB), Bernoulli Naive Bayes (BNB), Multinomial Naive Bayes (MNB) from the Naive Bayes category; Decision Tree Classifier (DTC), Extremely Randomized Tree Classifier (ERTC), and Nearest Centroid (NC) from the Decision Trees category; K-Nearest Neighbors (KNN); Ridge Regression (RR), Stochastic Gradient Descent (SGD), Passive Aggressive (PA), and Linear Perceptron (LP) from the Linear Models category; Linear Discriminant Analysis (LDA) from the Discriminant Analysis category; Support Vector Classification (SVC) and Nu-Support Vector Classification (Nu-SVC) from the Support Vector Machines category; Gaussian Process Classifier (GPC) from the Gaussian Processes category; Bagging Classifiers (BC), Random Forest Classifier (RFC), Extra Tree Classifier (ETC), Adaptive Boosting (ADA), Gradient Boosting Classifier (GBC), and Histogram-based Gradient Boosting (HGBT) from the Ensemble Methods category. These classifiers cover a broad range of approaches, from Naive Bayes and Decision Trees to Ensemble Methods and Support Vector Machines.

We evaluated the performance of 22 distinct classifiers. Table IX details the performance metrics, including accuracy, sensitivity, specificity, precision, f-score, and G-mean for each classifier, based on a 10-fold Cross Validation.

TABLE IX
Error enhancement for each improved algorithm. Each row shows that the corresponding algorithm.

| Category | Algorithm | Accuracy | Sensitivity | Specificity | Precision | F-SCORE | G-Mean |
|---|---|---|---|---|---|---|---|
| Ensemble Methods | BC | 0.73 | 0.73 | 0.87 | 0.72 | 0.72 | 0.64 |
| | ETC | 0.75 | 0.75 | 0.90 | 0.75 | 0.74 | 0.65 |
| | RFC | 0.74 | 0.74 | 0.91 | 0.73 | 0.71 | 0.61 |
| | HGBT | 0.76 | 0.76 | 0.88 | 0.76 | 0.75 | 0.69 |
| | ADA | 0.75 | 0.75 | 0.85 | 0.75 | 0.75 | 0.70 |
| | GBC | 0.78 | 0.78 | 0.89 | 0.77 | 0.77 | 0.70 |
| Discriminant Analysis | LDA | 0.75 | 0.75 | 0.86 | 0.74 | 0.74 | 0.69 |
| Gaussian Processes | GPC | 0.74 | 0.74 | 0.91 | 0.73 | 0.71 | 0.61 |
| Naive Bayes | GNB | 0.69 | 0.69 | 0.77 | 0.69 | 0.69 | 0.64 |
| | BNB | 0.56 | 0.56 | 0.54 | 0.61 | 0.57 | 0.57 |
| | CNB | 0.68 | 0.68 | 0.72 | 0.69 | 0.68 | 0.66 |
| | MNB | 0.71 | 0.71 | 0.84 | 0.70 | 0.70 | 0.62 |
| Linear Models | SGD | 0.73 | 0.73 | 0.84 | 0.74 | 0.72 | 0.65 |
| | RR | 0.76 | 0.76 | 0.88 | 0.76 | 0.75 | 0.69 |
| | PA | 0.67 | 0.67 | 0.78 | 0.70 | 0.65 | 0.55 |
| | LP | 0.63 | 0.63 | 0.59 | 0.71 | 0.61 | 0.59 |
| Decision Trees | NC | 0.64 | 0.64 | 0.66 | 0.67 | 0.65 | 0.63 |
| | ERTC | 0.64 | 0.64 | 0.72 | 0.64 | 0.64 | 0.59 |
| | DTC | 0.65 | 0.65 | 0.72 | 0.66 | 0.66 | 0.62 |
| Support Vector Machines | SVC | 0.76 | 0.76 | 0.91 | 0.76 | 0.74 | 0.65 |
| | Nu-SVC | 0.76 | 0.76 | 0.89 | 0.75 | 0.74 | 0.67 |
| K- Nearest Neighbors | KNN | 0.72 | 0.72 | 0.86 | 0.70 | 0.70 | 0.62 |

From the results, Gradient Boosting Classifier (GBC) from the ensemble methods category demonstrates the highest overall performance with an accuracy of 0.78, sensitivity of 0.78, and G-mean of 0.70. Notably, in terms of specificity, classifiers like the Gaussian Process Classifier (GPC) and Random Forest Classifier (RFC) perform exceptionally well with values over 0.90, indicating their ability to identify negative results correctly. Naive Bayes classifiers tend to show lower performance in our experiments, with the Bernoulli Naive Bayes (BNB) having the lowest accuracy, F-score, and G-mean among all tested classifiers.

However, while the comparison with other classifiers is informative, our study focuses on evaluating our proposed algorithms. When comparing the results of our proposed algorithms in Table VII with the results in Table IX, our proposed CODEL-CG-PR algorithm shows the most promising performance. With an average accuracy of 79.21, a sensitivity of 79.79, and a G-mean of 79.00, CODEL-CG-PR outperforms all other tested classifiers, including the well-performing GBC. It is particularly notable that CODEL-CG-PR achieves the best rank in accuracy, precision, F-score, and G-mean, substantiating the strength of this algorithm in balanced and high-performing classification.

## 5. Conclusion

Our research proposes an enhanced training algorithm for ECG signal classification using differential evolution (DE), opposition-based learning, and a region-based strategy. The algorithm starts by extracting features such as BPM, IBI, and SDNN from the ECG signals, which were then used as the inputs for a neural network. The weights of the multi-layer neural networks were updated using the DE algorithm, a region-based strategy, and opposition-based learning. The region-based strategy employed a clustering algorithm to identify good solutions in the current population, while the opposition-based learning aimed to improve the DE algorithm's exploration capabilities. Subsequently, the weights determined by the enhanced DE algorithm were used to initialize six different gradient-based local search algorithms. Our experimental results confirmed the superior performance of the proposed approach compared to the baseline algorithm.

Despite the satisfactory performance of the proposed algorithm, the authors intend to extend this work as follows.

• In this paper, we only optimize the training process. In the future, the optimization of architecture is under investigation.

• Another direction is to extend this work as a multi-objective optimization

• The performance of this research can be extended by some concepts, such as chaotic variables.

• The proposed DE algorithm can be used in other applications such as image segmentation.

## Acknowledgment

This work is funded by FCT/MCTES through national funds and when applicable co-funded EU funds under the project UIDB/EEA/50008/2020 (Este trabalho é financiado pela FCT/MCTES através de fundos nacionais e quando aplicável cofinanciado por fundos comunitários no âmbito do projeto UIDB/EEA/50008/2020).

This article/publication is based on work from COST Action IC1303 - AAPELE - Architectures. Algorithms and Protocols for Enhanced Living Environments and COST Action CA16226 - SHELD-ON - Indoor living space improvement: Smart Habitat for the Elderly. Supported by COST (European Cooperation in Science and Technology). More information on www.cost.eu.

Furthermore, we would like to express our sincere gratitude for the support and funding received from multiple sources. Firstly, we acknowledge the Centro Regional Operational Program (Centro 2020) for their contribution within the scope of research activities of Project CENTRO-01-0145-FEDER-000019-C4-Cloud Computing Competence Center. Secondly, we appreciate the fellowship contract provided within the framework of research activities PHArA-ON, funded by the European Union's Horizon 2020 research and innovation program. These scholarships and fellowships have greatly contributed to our research and made this work possible.